\documentclass[runningheads]{llncs}

\usepackage[mobile]{eccv}

\usepackage{eccvabbrv}

\usepackage{graphicx}
\usepackage{booktabs}

\usepackage[accsupp]{axessibility}  %

\usepackage{hyperref}

\usepackage{orcidlink}
\usepackage{amsmath}
\usepackage{amssymb}
\usepackage{bbold}
\usepackage{bbm}
\usepackage{dsfont}
\usepackage{graphicx}
\usepackage{lscape}
\usepackage{tabularx}
\usepackage{makecell}
\usepackage{multirow,bigstrut}
\usepackage{longtable}
\usepackage{ragged2e}

\usepackage[dvipsnames]{xcolor}

\newcommand{\projname}[0]{\textsc{PartSTAD}}

\definecolor{color_1}{RGB}{255,0,128}
\definecolor{color_2}{RGB}{128,128,0}

\definecolor{color_green}{RGB}{0, 168, 0}
\definecolor{color_lightgreen}{RGB}{112, 229, 0}
\definecolor{color_cyan}{RGB}{0, 236, 116}
\definecolor{color_yellow}{RGB}{255, 196, 0}
\definecolor{color_orange}{RGB}{255, 114, 32}
\definecolor{color_skyblue}{RGB}{0, 210, 255}
\definecolor{color_deepblue}{RGB}{0, 82, 255}

\renewcommand{\figurename}{Figure}

\begin{document}

\title{PartSTAD: 2D-to-3D Part Segmentation Task Adaptation}

\author{Hyunjin Kim\inst{1}\textsuperscript{\textdagger}\orcidlink{0009-0003-0922-6401} \and
Minhyuk Sung\inst{2}\orcidlink{0000-0001-7428-9570}
}
\authorrunning{H. Kim and M. Sung}

\institute{KRAFTON Inc., South Korea
\and
KAIST, South Korea \\
\email{rlaguswls98@krafton.com},~ \email{mhsung@kaist.ac.kr}}

\def\thefootnote{\textdagger}\footnotetext{This work was conducted when the author was at KAIST.}\def\thefootnote{\arabic{footnote}}

\maketitle
{
    \centering
    \captionsetup{type=figure}
    \includegraphics[width=\textwidth]{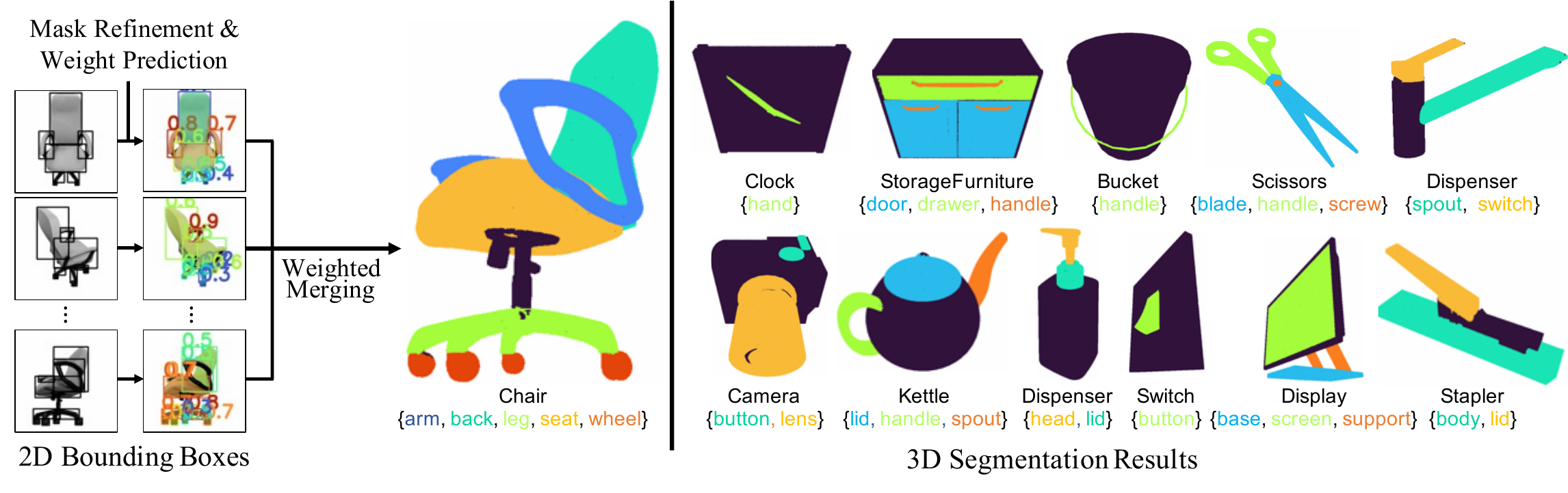}
    \captionof{figure}{We introduce \projname{}, a novel few-shot 3D point cloud part segmentation method that leverages 2D-to-3D task adaptation. By obtaining 2D segmentation masks in multi-view images from GLIP~\cite{li2022glip} and SAM~\cite{kirillov2023segment} and optimizing the mask weights for 3D segmentation as a learning objective, it can successfully predict fine-grained parts with accurate boundaries, as shown in the figure above.}
    \label{fig:teaser}
}

\begin{abstract}
We introduce~\projname{}, a method designed for the task adaptation of 2D-to-3D segmentation lifting. Recent studies have highlighted the advantages of utilizing 2D segmentation models to achieve high-quality 3D segmentation through few-shot adaptation. However, previous approaches have focused on adapting 2D segmentation models for domain shift to rendered images and synthetic text descriptions, rather than optimizing the model specifically for 3D segmentation. Our proposed task adaptation method finetunes a 2D bounding box prediction model with an objective function for 3D segmentation. We introduce weights for 2D bounding boxes for adaptive merging and learn the weights using a small additional neural network. Additionally, we incorporate SAM, a foreground segmentation model on a bounding box, to improve the boundaries of 2D segments and consequently those of 3D segmentation. Our experiments on the PartNet-Mobility dataset show significant improvements with our task adaptation approach, achieving a \textbf{7.0\%p} increase in mIoU and a \textbf{5.2\%p} improvement in mAP\textsubscript{50} for semantic and instance segmentation compared to the SotA few-shot 3D segmentation model. The code is available at \url{https://github.com/KAIST-Visual-AI-Group/PartSTAD}.


\end{abstract}
    
\section{Introduction}
\label{sec:intro}
3D segmentation has been a subject of extensive research in computer vision due to its central role in various downstream applications involving the understanding of shape structure, functionality, mobility, and semantics. However, the limited availability of annotated 3D shapes has remained a bottleneck in achieving generalizability in learned segmentation models for diverse 3D data. Annotating 3D segmentation is particularly labor-intensive, time-consuming, and requires expertise in handling 3D models. For this reason, the scale of 3D part-level annotations remains in the tens of thousands~\cite{Mo_2019_partnet}, while the scale of 2D annotation datasets in the image domain, for example, exceeds a million~\cite{lin2014microsoftcoco,kirillov2023segment}.

Recent research~\cite{abdelreheem2023satr,liu2023partslip} has illuminated the potential of visual-language models bridging textual descriptions with images in accomplishing zero-shot or few-shot 3D segmentation results. The basic idea is to render a 3D model from various viewpoints, conduct 2D detection or segmentation on the rendered images, and then aggregate the 2D segmentation results into a 3D representation using either a voting mechanism~\cite{liu2023partslip} or a label propagation scheme~\cite{abdelreheem2023satr}. This approach is not only effective in enabling zero-shot and few-shot 3D segmentation but also offers an advantage in that the set of part names does not need to be predefined in training but can be determined at test time.

PartSLIP~\cite{liu2023partslip} is a notable example that achieves 3D part segmentation results comparable to fully supervised methods while adapting a pretrained 2D detection model with few-shot training for 3D segmentation. It leverages the pretrained GLIP~\cite{li2022glip} model for 2D detection, associating the output 2D bounding boxes with one of the tokens (part names) provided in the input prompt. Its key component is the GLIP finetuning process, which introduces a small number of learnable parameters to the frozen GLIP model and trains them using few-shot synthetic images and texts employed in the 3D segmentation pipeline. As GLIP was initially trained with real photos and natural language descriptions of objects by humans, this adaptation to rendered images and the unconventional description (a sequence of part names) results in a substantial improvement in the 3D segmentation task.

Despite promising initiatives, it is important to note that the previous approach was limited to achieving \emph{domain adaptation}. In contrast, we emphasize that in the 2D-to-3D segmentation lifting task, not only does the data domain change, but the task itself also shifts from 2D segmentation to 3D segmentation. Thus, the process of applying the pretrained model to the new task requires \emph{task adaptation}, involving modifying the model with an objective function associated with the new task. Particularly for 3D segmentation, it is necessary to integrate 2D segmentation results from multiple viewpoints into a coherent 3D representation. Therefore, it is crucial to control the noise of 2D bounding box predictions in the context of its impact on the final 3D segmentation after integration.

Specifically, our Part Segmentation Task ADaptation method,~\projname{}, adapts the pretrained GLIP model with a relaxed 3D mIoU loss while incorporating bounding boxes across all different views. Instead of typical finetuning, we draw inspiration from recent work~\cite{liu2023partslip,hu2021lora} that introduces small learnable parameters to the existing model while keeping existing parameters frozen. As a result, we train a small MLP using 2D bounding box features extracted from the pretrained GLIP model. Since the relaxed mIoU loss is non-differentiable with respect to the bounding box positions, we suggest predicting a \emph{weight} for each bounding box instead of adjusting its position. This results in minimal modification in the 2D-to-3D voting scheme while achieving a significant improvement in 3D segmentation, even with few-shot training, e.g., with eight objects per category, as done in PartSLIP. To further enhance 3D segmentation performance, we leverage SAM~\cite{kirillov2023segment}, a 2D instance segmentation model, to segment the foreground region within each bounding box, obtaining an accurate boundary for each 2D segment.

In our experiments on the PartNet-Mobility~\cite{Xiang_2020_SAPIEN} dataset, we showcase the superior performance of our method in comparison to recent zero-shot/few-shot 3D semantic and instance segmentation methods that leverage 2D segmentation models. In contrast to the SotA few-shot method, our approach achieves a \textbf{7.0\%p} improvement in semantic segmentation mIoU and a \textbf{5.2\%p} improvement in instance segmentation mAP\textsubscript{50}. These improvements are consistent across all object classes.

\section{Related Work}
\subsection{Supervised 3D Segmentation}

Given the availability of open segment-annotated 3D datasets~\cite{armeni2017s3dis, hackel2017semantic3d, behley2019semantickitti, dai2017scannet, chang2015shapenet, Mo_2019_partnet, Xiang_2020_SAPIEN, Matterport3D}, 3D segmentation has been extensively researched in the last several years, focusing on the development of novel network architectures for supervised learning. Regarding the architecture of semantic segmentation, diverse models have been explored, including PointNet~\cite{qi2017pointnet} and its variants~\cite{qi2017pointnet++,ma2022pointmlp,qian2022pointnext,zhang2023parameter,Xiang_2021_ICCV_curvenet}, those using CNN~\cite{li2018pointcnn,thomas2019kpconv,xu2021paconv,liu2019densepoint}, GCN~\cite{huang2020lsgcn,dgcnn,li2019tgnet}, and Transformers~\cite{yan2020pointasnl,zhao2021pointtransformer,wu2022pointtransformerv2,yu2021pointbert}. For instance segmentation, various approaches based on proposing 3D bounding boxes~\cite{hou20193dsis, yi2019gspn, yang20193dbonet}, clustering learned features~\cite{wang2018sgpn, narita2019panopticfusion, jiang2020pointgroup, vu2022softgroup, Zhang_2021_CVPR,  liu2020selfpred, zhao2021prototype, chu2021icm3d}, and combining both semantic and instance segmentations~\cite{wang2019asis, pham-jsis3d-cvpr19, liang20203dgel,Schult23ICRA_mask3d} have been introduced. Despite these great advances, the performance of supervised methods has been limited by the scale of 3D datasets; the largest part-annotated 3D dataset, PartNet~\cite{Mo_2019_partnet}, includes fewer than 30k models.

There have been some attempts to overcome the limitation by leveraging additional weak supervisions, such as text descriptions of objects~\cite{koo2022partglot}, 3D bounding box annotations~\cite{chibane2021box2mask}, sparse labels~\cite{Li_2022_CVPR_hybridcr, liu2021oneclick}, and geometric priors~\cite{li2019spfn, Le_2021_ICCV_cpfn}. However, the scale of this additional data has also been limited to tens of thousands, while datasets in the image domain are on the scale of millions.

\subsection{3D Segmentation Using Vision-Language Models}

Recent work introduced ideas about exploiting vision-language models (VLM) for 3D segmentation. The advantage of utilizing the learned visual-language grounding in 3D segmentation lies in the strong generalizability to arbitrary 3D models with zero-shot or few-shot training, and also in the open vocabulary understanding that enables labeling points without specifying the set of part or object names during training. For a VLM, CLIP~\cite{radford2021clip} has been adapted in multiple previous works to lift the 2D grounding to the 3D, detecting parts in an object~\cite{decatur2023highlighter} and objects in a scene~\cite{peng2023openscene, takmaz2023openmask3d, chen2023clip2scene, zhang2023clipfo3d}, while the results are limited to highlighting some regions without providing clear segment boundaries.

To obtain precise segments in 3D, other 2D object detection and segmentation models, such as SAM~\cite{kirillov2023segment}, GLIP~\cite{li2022glip}, and GroundingDINO~\cite{liu2023groundingdino}, are also leveraged. SAM3D~\cite{yang2023sam3d} was an example of directly lifting the 2D masks from SAM to 3D with a bottom-up merging approach, although it was limited to segmenting the parts, not labeling them. SA3D~\cite{cen2023sa3d} utilized NeRF representation and allowed user interaction for 3D segmentation, but it can be inefficient when the 3D objects are given as a mesh, point cloud, or other common 3D representations. OpenMask3D~\cite{takmaz2023openmask3d} and OpenIns3D~\cite{huang2023openins3d} are concurrent works, both of which combine multiple foundation models, SAM~\cite{kirillov2023segment} and CLIP~\cite{radford2021clip} for OpenMask3D, and GroundingDINO~\cite{liu2023groundingdino} and LISA~\cite{lai2023lisa} for OpenIns3D. These works focus on segmenting objects in 3D scenes, while we aim to do finer-grained segmentation, finding parts in 3D objects.

PartSLIP~\cite{liu2023partslip} and SATR~\cite{abdelreheem2023satr} are notable examples that achieve part segmentation of 3D objects based on a 2D object detection model, GLIP~\cite{li2022glip}. They obtain 2D bounding boxes on the rendered images from multiple views and integrate them over the 3D object with different merging schemes. The finetuning of GLIP introduced by PartSLIP particularly leads to a performance improvement, but it is limited to domain adaptation, not task adaptation. Building on this, we introduce a novel task adaptation technique as well as integration with SAM~\cite{kirillov2023segment} to achieve a further significant improvement in 3D segmentation accuracy.

\subsection{Task Adaptation}
Taskonomy~\cite{taskonomy2018} is a seminal work that introduced the concept of task transfer learning for the first time, enabling the transformation of an image dense prediction model trained for a base task to perform other tasks through finetuning. Pruksachatkun \etal.~\cite{pruksachatkun-etal-2020-intermediate} also introduced a similar idea for language-domain tasks. Later, task adaptation has been extensively studied to leverage features learned from generative models, such as GAN~\cite{zhang2021-datasetgan} and diffusion models~\cite{baranchuk2022labelefficient, xu2023odise, Tang2023Emergent, yang2023dmrl}, adapting them with a small network for various tasks, including image segmentation~\cite{zhang2021-datasetgan, baranchuk2022labelefficient,xu2023odise}, depth prediction~\cite{Tang2023Emergent}, and keypoint detection~\cite{yang2023dmrl}. For 3D tasks, Abdelreheem \etal.~\cite{abdelreheem2023zero} are notable examples of using vision language-grounding models for 3D correspondence but are limited to directly applying 2D priors without finetuning. We introduce a novel task adaptation technique that lifts 2D segments to 3D by adapting the 2D features during the training of a small network with a task-specific objective function.

\section{Background and Preliminaries}
Recent research~\cite{liu2023partslip,abdelreheem2023satr,decatur2023highlighter} has demonstrated how vision-language models designed for image object detection and segmentation can be applied to segment parts in 3D objects. The main idea involves rendering a 3D model from various viewpoints, performing 2D detection or segmentation for the rendered image at each view, and then combining the 2D segmentation results over the 3D model.

Specifically, SATR~\cite{abdelreheem2023satr} utilizes GLIP~\cite{li2022glip} as the 2D detection model. To obtain 2D bounding boxes for each of the specified part names, it feeds a concatenated list of these part names (separated by commas) as the input prompt to the pretrained GLIP model, along with a rendered image. The 2D bounding boxes are then mapped to a single face of the input 3D mesh, located at the center of the respective 2D bounding box, and the part names are propagated throughout the entire mesh based on geodesic distances. Similarly, PartSLIP~\cite{liu2023partslip} also leverages GLIP~\cite{li2022glip} for 2D detection but employs a different voting scheme for the 2D bounding box integration over the 3D object, which utilizes the over-segmentation of the input point cloud. One key distinction between PartSLIP and SATR is that PartSLIP proposes to modify the GLIP model for domain adaptation to the rendered images and text prompts used in 3D segmentation. Specifically, the text prompt is given as a concatenation of part names in the PartSLIP pipeline, which is not a typical and natural description of the object. Therefore, it learns a category-specific constant offset vector for the language embedding feature in the given pairs of prompts and rendered images, while keeping the GLIP parameters frozen.

Our work builds upon the PartSLIP pipeline. In the following subsections, we describe the details of its major components: the voting scheme, and the finetuning process.

\subsection{PartSLIP~\cite{liu2023partslip}}

The main technical components of PartSLIP are the 1) voting scheme based on super points and 2) adaptation to text prompts\footnote[1]{Although multi-view feature aggregation is also introduced, the improvement by this component is marginal, and the author of PartSLIP also did not include this component in their official code. Therefore, we will omit this part in our review.}. We describe the details of each component below.

\paragraph{\textbf{Voting for Super Points.}}
PartSLIP first over-segments the input 3D object represented as a point cloud $\mathcal{P}$ into a set of super points $P_i \subseteq \mathcal{P}$, which provide geometric priors for the boundaries of the parts. The 3D object is rendered into $K$ viewpoints, and the 2D bounding boxes for each image are predicted using GLIP~\cite{li2022glip}. Let $\mathcal{B}$ denote the entire set of 2D bounding boxes from all views, and $B_{jk} \subseteq \mathcal{B}$ indicates a subset that includes the bounding boxes classified as the $j$-th part label from the $k$-th view. $\mathrm{V}_{k}: \mathcal{P} \rightarrow \{0, 1\}$ is a function indicating whether the input point is visible from the $k$-th viewpoint. Also, for each 2D bounding box $b \in \mathcal{B}$, $\mathrm{I}_b: \mathcal{P} \rightarrow \{0, 1\}$ is a function indicating whether the input point is included in the bounding box $b$. Given these, the voting aggregating the semantic labels of the 2D bounding boxes to the 3D point cloud is performed by calculating the following score $s_{ij}$ for each pair of the $i$-th super point and the $j$-th label:
\begin{align}
s_{ij}=\frac{\sum_k\sum_{p \in P_{i}} \mathrm{V}_k(p) \left( \max_{b \in (B_{jk} \cup  \{b_{\emptyset}\})} \mathrm{I}_b(p) \right)}{\sum_k\sum_{p \in P_{i}} \mathrm{V}_k(p)},
\label{eq:score}
\end{align}
where $b_{\emptyset}$ is a \emph{null} bounding box, and we assume that $\mathrm{I}_{b_{\emptyset}}$ for the null bounding box returns zero for any input point (simplifying notations for the case when $B_{jk} = \emptyset$). Note that this $s_{ij} \in [0, 1]$ denotes the ratio of visible points from each view included in any of the 2D bounding boxes labeled with the $j$-th label. 
Finally, for each $i$-th super point, the label with the highest score $s_{ij}$ across all the labels is assigned to the super point. Exceptionally, if the highest score is less than a null label threshold $s_{\emptyset}$, which is set to 0.5 in our experiments, the null label is assigned to the super point.

\paragraph{\textbf{GLIP Finetuning.}}
While GLIP demonstrates impressive generalizability for unseen images and prompts, it still has limitations when it comes to handling synthetic images and text prompts, such as the rendered images and sequences of part names used in the PartSLIP pipeline. To address this, the authors of PartSLIP propose adding small, learnable parameters to the GLIP model while keeping all the pretrained GLIP parameters frozen. These new parameters are learned with very few-shot examples: 8 annotated 3D objects per category. They represent offset feature vectors for each part name and remain constant as global variables for each category, rather than changing for each input. This adaptation of GLIP significantly improves the accuracy of 3D part segmentation.%

\paragraph{\textbf{Instance Segmentation.}}
The voting scheme described above is designed for semantic segmentation, but PartSLIP also introduces a straightforward merging-based method to achieve instance segmentation based on semantic segmentation. To obtain instance segments, it merges adjacent super points under two conditions: 1) they share the same semantic label, and 2) they are either both included or both excluded for all bounding boxes.

\begin{figure*}[!ht]
\begin{center}
\includegraphics[width=\textwidth]{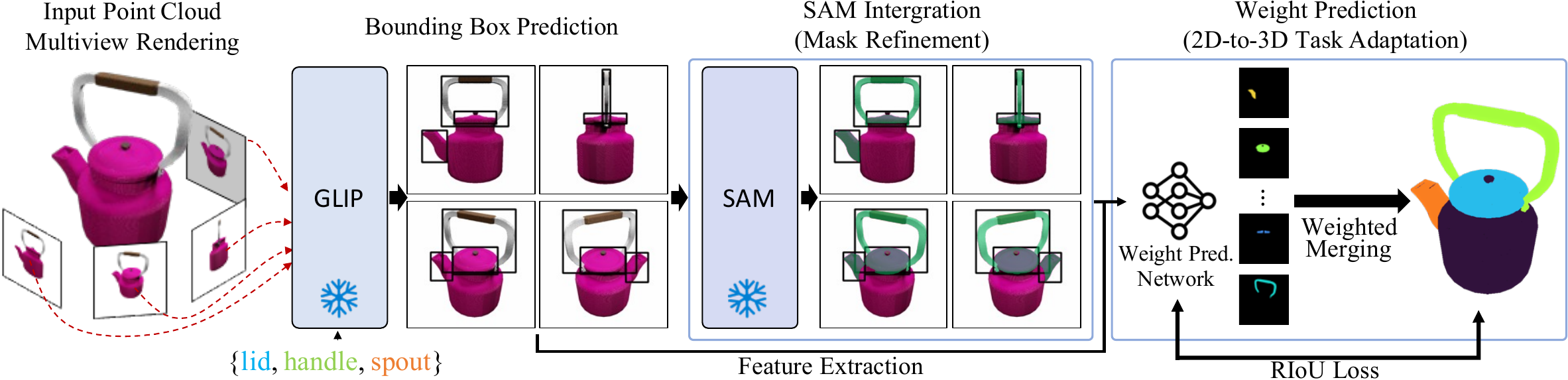} 
\end{center}
\caption{\textbf{Overall pipeline of~\projname{}}.
Our approach begins by rendering the provided 3D point cloud from multiple viewpoints. Subsequently, we extract 2D bounding boxes for its parts using GLIP~\cite{li2022glip} (Bounding Box Prediction); note that we utilize the finetuned GLIP model from PartSLIP~\cite{liu2023partslip}. Following this, we convert the bounding boxes into segmentation masks using SAM~\cite{kirillov2023segment}, extracting the foreground region for each bounding box (SAM Mask Integration). Next, we predict weights for all the masks and adaptively combine them into a 3D representation (2D-to-3D task adaptation). The final step involves obtaining the segmentation label for the input point cloud. The GLIP and SAM models are frozen, while only our novel weight prediction network is trained per category in a few-shot setting (8 objects).}

\label{fig:overall_pipeline}
\end{figure*}

\section{\projname{}: Task Adaptation for 2D-to-3D}

PartSLIP has demonstrated that adapting a 2D detection model for 3D segmentation with a small set of trainable parameters can significantly enhance 3D segmentation performance. However, we observe that PartSLIP's adaptation is limited to \emph{domain adaptation}, where it learns new parameters using synthetic images and data used in the PartSLIP framework without altering the \emph{objective function} during training.

When applying the 2D segmentation model to 3D segmentation, it is important to note that not only does the data domain change, but the \emph{task} itself changes. Therefore, through the process of incorporating 2D bounding boxes into the 3D object, the 2D bounding box prediction model must align with our ultimate goal: 3D part segmentation.

To address this, we introduce Part Segmentation Task ADaptation method,~\projname{}, a \emph{task adaptation} approach designed to finetune the 2D segmentation model for 3D part segmentation. In Sec.~\ref{sec:mriou}, we introduce our objective function to finetune the GLIP for the 3D segmentation task. In Sec.~\ref{sec:weight_prediction}, we describe how we adapt the pretrained GLIP model with the new objective function by introducing additional learnable parameters and modifying the voting scheme (Eq.~\ref{eq:score}). PartSLIP also faces a performance limitation due to the GLIP that provides 2D bounding boxes, not 2D segments. In Sec.~\ref{sec:sam_mask}, we also propose to combine the SAM\cite{kirillov2023segment} foreground mask for each bounding box to obtain more precise 2D segments and, consequently, further improve the 3D segmentation.

\subsection{3D mRIoU Loss}
\label{sec:mriou}
We begin by describing the loss function that we use for adapting the GLIP to the 3D segmentation task. Let $\mathbf{l}_j$ and $\hat{\mathbf{l}}_j \in \{0,1\}^{|\mathcal{P}|}$ be binary vectors indicating whether each point $p$ in $\mathcal{P}$ has the $j$-th label in the ground truth and label prediction, respectively. The mean Intersection over Union (mIoU) for the predicted 3D segmentation, the standard evaluation metric for 3D segmentation, is defined as follows:
\begin{align}
\text{mIoU} (\{\mathbf{l}_j\}, \{\hat{\mathbf{l}}_j\}) = \frac{1}{M} \sum_{j = 1}^{M} \frac{\mathbf{l}_j^\top \hat{\mathbf{l}}_j}{\|\mathbf{l}_j\|_1 + \| \hat{\mathbf{l}}_j \|_1 - \mathbf{l}_j^\top \hat{\mathbf{l}}_j},
\end{align}
where $M$ is the number of labels.

Our goal is to use the 3D mIoU as the objective function directly when adapting the GLIP model. However, the mIoU is non-differentiable. Thus, we employ a relaxed version known as mean Relaxed IoU (mRIoU)~\cite{deeppartinduction, li2019spfn}, which simply allows the predicted indicator vectors $\hat{\mathbf{l}}_j$ to be non-binary: $\hat{\mathbf{l}}_j \in [0,1]^{|\mathcal{P}|}$. The loss $\mathcal{L}_{\text{mRIoU}}$ is specifically defined as follows:
\begin{align}
\mathcal{L}_{\text{mRIoU}} = 1 - \text{mIoU} (\{\mathbf{l}_j\}, \{\hat{\mathbf{l}}_j\}).
\label{eq:mriou}
\end{align}

In the supplementary material, we demonstrate that the choice of the loss function, as compared to other alternatives such as the cross-entropy loss, is crucial to achieving a substantial improvement in our task adaptation.

Note that in the PartSLIP framework, where super points are utilized in 3D segmentation, and $s_{ij} \in [0, 1]$ indicates the likelihood of the $i$-th super point belonging to the $j$-label, the mRIoU can be calculated by defining $\hat{\mathbf{l}}_j$ based on $\mathbf{s}_j = [s_{1j}, s_{2j}, \cdots]^\top$ as follows:

\begin{align}
\hat{\mathbf{l}}_j = \mathbf{M} \mathbf{s}_j,
\end{align}

where $\mathbf{M} \in \{0, 1\}^{|\mathcal{P}| \times | \{P_i\} |}$ is a binary matrix that describes the memberships from each point to the super points.

\subsection{Bounding Box Weight Prediction}
\label{sec:weight_prediction}
A typical approach for adapting a pretrained model to a new task would be finetuning the model using an objective function tailored to that new task. However, the finetuning method generally fails to yield meaningful improvements, especially when the goal is to achieve few-shot adaptation (e.g., with only 8 annotated 3D objects per class, as done in our case), while the pretrained model has been trained on an extensive dataset. To address this challenge, recent works like LoRA~\cite{hu2021lora} and PartSLIP~\cite{liu2023partslip} have introduced the concept of adding small learnable parameters while keeping the existing pretrained model parameters frozen. This approach not only makes training very efficient but also enables the model to generalize effectively to unseen data.

While we also follow this approach, our specific challenge arises from the fact that when we use the 3D mRIoU (Eq.~\ref{eq:mriou}) as the objective function for 3D segmentation, it becomes non-differentiable with respect to the 2D bounding box location, which is the output of GLIP. This non-differentiability arises from the computation of the score $s_{ij}$ (Eq.~\ref{eq:score}) for each pair of super point and label.

To address this, we propose training a small network that does not refine the location but instead predicts a \emph{weight} for each 2D bounding box, taking the learned features of the boxes as input.
If we denote the output weight of the network for bounding box $b$ as $\mathrm{W}(b) \in \mathbb{R}^{+}$, 
the score $s_{ij}$ for a pair of super point and label is modified to $\tilde{s}_{ij}$ as follows:
\begin{align}
\tilde{s}_{ij} = \frac{\sum_k\sum_{p \in P_{i}} \mathrm{V}_k(p) \left( \max_{b \in (B_{jk} \cup  \{b_{\emptyset}\})} \mathrm{I}_b(p) \mathrm{W}(b) \right)}{\sum_k\sum_{p \in P_{i}} \mathrm{V}_k(p)}.
\end{align}
Note that the change is simply to multiply the weight $\mathrm{W}(b)$ by $\mathrm{I}_b(p)$, while only $\mathrm{I}_b(p)$, indicating whether the point $p$ is included in bounding box $b$, has been used in Eq.~\ref{eq:score}.
This modified score no longer falls within the range of $[0, 1]$. The final score $\bar{s}_{ij}$ is thus defined by normalizing $\tilde{s}_{ij}$ using the softmax function over the set of labels:
\begin{align}
\bar{s}_{ij} = \frac{\exp(\tilde{s}_{ij})}{\sum_j \exp(\tilde{s}_{ij})}.
\end{align}

We additionally propose to make the unnormalized score for the null label $\tilde{s}_{\emptyset}$ \emph{learnable} in our case, initialized with 10. We thus compute the softmax above while including $\tilde{s}_{\emptyset}$. The label of each super point is still chosen as the label giving the highest score $\bar{s}_{ij}$, becoming null if the normalized null label score $\bar{s}_{\emptyset}$ becomes the highest.

\paragraph{\textbf{Network Architecture.}} We design the weight predictor network to take the bounding box feature vectors $\mathbf{f}_b$ from the pretrained GLIP model as input. The feature vectors of all bounding boxes across all views are fed to the network at once and processed with a small shared two-layer MLP. Context normalization~\cite{yi2018learningCNe} is added in the middle of the two layers to incorporate contextual information from the global set of boxes for each box. The output of the MLP for each bounding box is further processed with the following modified ReLU function $\phi(\cdot)$:
\begin{align}
\phi(\mathbf{x}) = \max\left(\tau + \mathbf{x}, 0 \right),
\end{align}
where $\tau$ is a user-defined constant offset (10 in our experiments).

\subsection{SAM~\cite{kirillov2023segment} Mask Integration}
\label{sec:sam_mask}
Another limitation of PartSLIP is its reliance on GLIP~\cite{li2022glip} for 2D segmentation. Since GLIP is, in turn, a 2D object detection model, it produces bounding boxes instead of 2D segments, which cannot provide accurate boundaries of segments. We propose to address this issue by incorporating another pretrained 2D segmentation model, SAM~\cite{kirillov2023segment}. SAM has the functionality of taking a bounding box as input and producing the foreground mask. Using this, we replace the set of 2D bounding boxes $\mathcal{B}$ with a set of 2D masks, while preserving the remaining steps in the framework. Note that the point-to-bounding-box membership $\mathrm{I}_b$ is changed to point-to-mask membership, and we still use the same bounding box features extracted from GLIP corresponding to each mask to train the weight prediction network. We demonstrate the effectiveness of applying the SAM mask in our experiments in Sec.~\ref{sec:results}.

\begin{figure*}[t!]
    \centering
    \scriptsize{
        \renewcommand{\arraystretch}{0.0}
        \setlength{\tabcolsep}{0.0em}
        \setlength{\fboxrule}{0.0pt}
        \setlength{\fboxsep}{0pt}
        
        \begin{tabularx}{\textwidth}{>{\centering\arraybackslash}m{0.15\textwidth}>{\centering\arraybackslash}m{0.85\textwidth}}\\
         & \framebox{\includegraphics[width=0.85\textwidth]{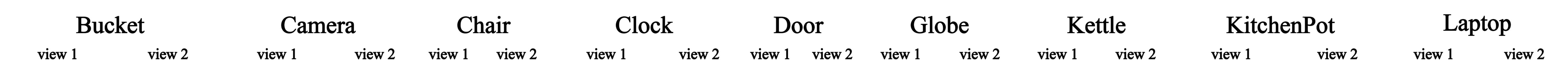}} \\
        {\tiny\makecell[c]{Input}} & \framebox{\includegraphics[width=0.85\textwidth]{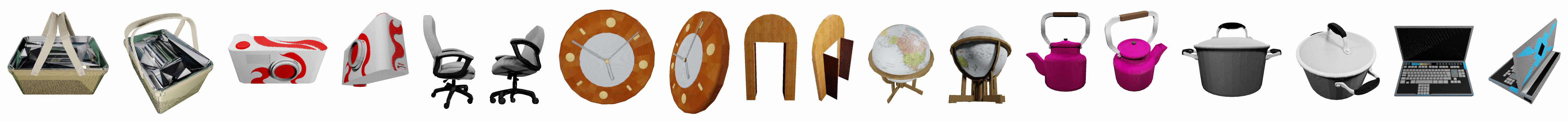}} \\
        {\tiny\makecell[c]{GT}} & \framebox{\includegraphics[width=0.85\textwidth]{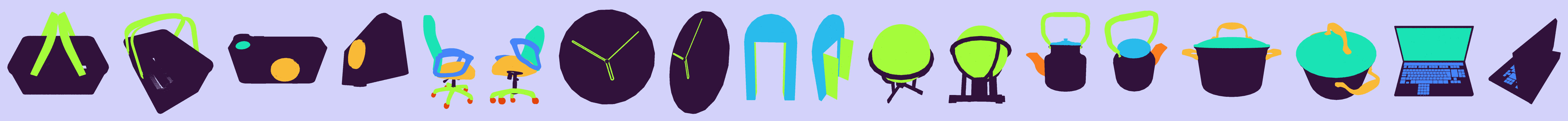}} \\
        {\tiny\makecell[c]{SATR~\cite{abdelreheem2023satr}}} & \framebox{\includegraphics[width=0.85\textwidth]{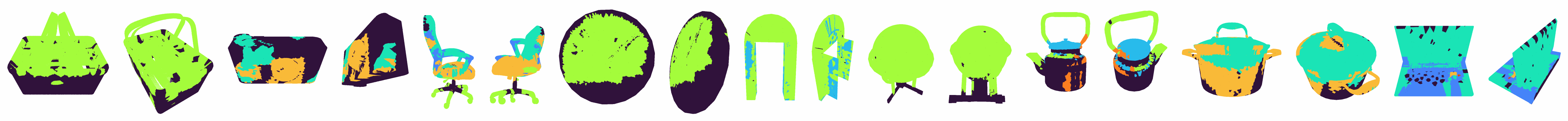}} \\
        {\tiny\makecell[c]{SATR~\cite{abdelreheem2023satr}+SP}} & \framebox{\includegraphics[width=0.85\textwidth]{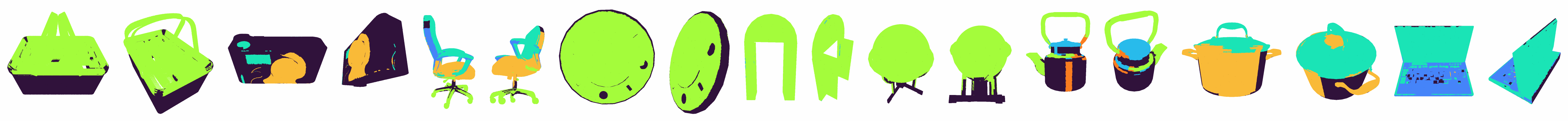}} \\
        {\tiny\makecell[c]{PartSLIP~\cite{liu2023partslip}}} & \framebox{\includegraphics[width=0.85\textwidth]{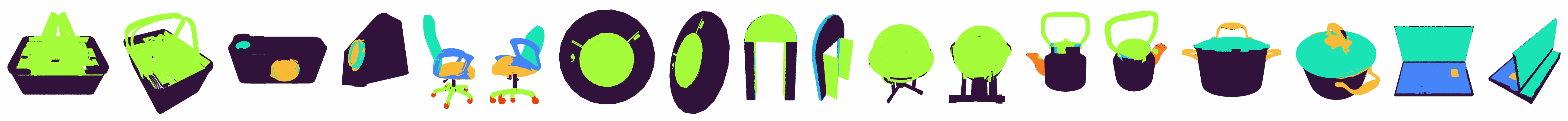}} \\ 
        {\tiny\makecell[c]{\projname{} \\(Ours)}} & \framebox{\includegraphics[width=0.85\textwidth]{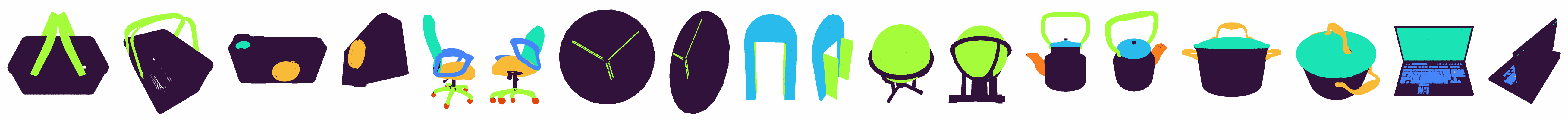}} \\
        \end{tabularx}
    }
    \caption{Qualitative comparison of semantic segmentation results. Our \projname{} segments 3D parts more precisely with clearer boundaries, even for small (Camera, Chair) and thin (Clock) parts.}
    \label{fig:semantic_qualitative}
\end{figure*}

\begin{table*}[t!]
    \caption{Quantitative comparison of the semantic segmentation results (mIoU(\%)) on the PartNet-Mobility dataset. Ours achieves a \textbf{7.0\%p} improvement in average mIoU compared to PartSLIP~\cite{liu2023partslip}, the SotA few-shot 3D segmentation method, while consistently increasing mIoU across all categories. Please refer to the supplementary material for the complete table with results for all 45 categories.}
    \centering
    {\scriptsize
    \setlength{\tabcolsep}{0.1em}
    \begin{tabularx}{\linewidth}{>{\centering\arraybackslash}m{0.2\textwidth}|>{\centering\arraybackslash}m{0.08\textwidth}|>{\centering\arraybackslash}m{0.09\textwidth}>{\centering\arraybackslash}m{0.07\textwidth}>{\centering\arraybackslash}m{0.07\textwidth}>{\centering\arraybackslash}m{0.07\textwidth}>{\centering\arraybackslash}m{0.07\textwidth}>{\centering\arraybackslash}m{0.07\textwidth}>{\centering\arraybackslash}m{0.07\textwidth}>{\centering\arraybackslash}m{0.07\textwidth}>{\centering\arraybackslash}m{0.07\textwidth}}\toprule
Method&mIoU&{\tiny\makecell[b]{Storage\\Furniture}}&{\tiny Table}&{\tiny Chair}&{\tiny Switch}&{\tiny Toilet}&{\tiny Laptop}&{\tiny USB}&{\tiny Remote}&{\tiny Scissors}\\
\midrule
SATR~\cite{abdelreheem2023satr}&29.3&20.6&23.3&33.1&21.4&17.6&11.2&30.2&17.2&36.8\\
SATR~\cite{abdelreheem2023satr}+SP&34.8&28.9&28.0&37.7&37.0&22.1&12.4&33.4&28.0&43.0\\
PartSLIP~\cite{liu2023partslip}&58.0&52.3&44.6&82.8&52.1&50.4&31.2&52.1&36.6&61.4\\
\midrule
\multicolumn{11}{c}{Ablation Study} \\
\midrule
\textit{w/o Weight Pred.}&61.9&56.6&45.0&85.0&51.9&56.6&31.5&57.1&36.6&60.8\\
\textit{w/o SAM Integ.}&62.1&54.0&45.7&83.1&53.0&49.4&33.6&53.6&46.8&67.5\\
\midrule
\projname{}    (Ours)&\textbf{65.0}&\textbf{59.5}&\textbf{47.8}&\textbf{85.3}&\textbf{57.9}&\textbf{57.5}&\textbf{34.6}&\textbf{59.9}&\textbf{53.4}&\textbf{68.5}\\
    \bottomrule
    \end{tabularx}
    \label{tab:semantic_quantitative}
    }
\end{table*}

\section{Experiment Results}
\label{sec:results}
In the experiments, we compare our method with the SotA 2D-prior-based zero-shot/few-shot 3D segmentation models using the PartNet-Mobility~\cite{Xiang_2020_SAPIEN} dataset and OmniObject3D~\cite{wu2023omniobject3d} dataset. In the \textbf{supplementary material}, we present additional results on the ablation study, outcomes with scanned data, and comprehensive results covering all PartNet-Mobility categories.

\subsection{Experiment Setup}
\paragraph{\textbf{Dataset.}}
We use the PartNet-Mobility~\cite{Xiang_2020_SAPIEN} dataset, which is also used in the experiments of PartSLIP~\cite{liu2023partslip}. The PartNet-Mobility~\cite{Xiang_2020_SAPIEN} dataset includes 45 object categories. For each category, the training split includes 8 shapes, while the test split has a range of the number of objects from 6 to 338, totaling 1,906 objects across all 45 categories.
Following PartSLIP, 8 shapes from the test set of each category are designated as the validation set.
To obtain the 2D bounding boxes for each object, we use 10 images from fixed viewpoints across all our experiments.
Note that the PartNet-Mobility~\cite{Xiang_2020_SAPIEN} dataset is identical to the subset of PartNet-Ensembled dataset used to train and evaluate PartSLIP~\cite{liu2023partslip}. (The remaining part of PartNet-Ensembled was only used to train other supervised baseline models, not PartSLIP~\cite{liu2023partslip}.)

\paragraph{\textbf{Evaluation Metrics.}}
As evaluation metrics, we employ the \textit{mean Intersection over Union} (\textbf{mIoU}) metric for semantic segmentation and \textit{mean Average Precision} (\textbf{mAP}) for instance segmentation.
 
For semantic segmentation, the average mIoU over the entire dataset is computed by first calculating it per semantic part, averaging them across the parts in the same category, resulting in category-level mIoU, and then averaging the category-level mIoUs again across all the categories.

For instance segmentation, we compute two types of mAP: \textit{part-aware} mAP and \textit{part-agnostic} mAP. For part-aware mAP, similar to mIoU in semantic segmentation, we first calculate the mAP for each semantic part with instances having the part label. We then average them for each category, and then across categories.
For part-agnostic mAP, we do not consider the semantic labels and directly compute the category-level mAP with part instances in the same object category, and then average them. All mAP values correspond to mAP$_{50}$ with an IoU threshold of 50\%.

\paragraph{\textbf{Baselines.}}
We compare our method with three baselines: PartSLIP~\cite{liu2023partslip}, SATR~\cite{abdelreheem2023satr}, and SAM3D~\cite{yang2023sam3d}. 

\begin{itemize}
    \item \textbf{PartSLIP}~\cite{liu2023partslip} is the method that we adopt as our base. Note that PartSLIP uses a finetuned GLIP model, and we also employ the same model in our framework to extract bounding box features for our task adaptation.
    \item \textbf{SATR} \cite{abdelreheem2023satr} is similar to PartSLIP, conducting 3D segmentation using 2D bounding boxes from GLIP~\cite{li2022glip}. However, SATR differs from PartSLIP in four key aspects: 1) it does not involve finetuning the GLIP model, 2) it does not utilize super points, 3) it is designed for mesh segmentation, not point cloud segmentation, and 4) it employs a distinct approach for integrating 2D bounding boxes into 3D (label propagation).
    To ensure a fair comparison between our method and SATR, we implement the following changes to SATR. First, since we employ the GLIP model finetuned by PartSLIP in our framework, we use the same finetuned model within the SATR framework. Second, we convert the point cloud input to a mesh, performing mesh segmentation with SATR, and apply the results to the input point cloud by finding the closest vertex for each point. Lastly, we introduce a version of SATR using super points, resulting in a comparison between our method and two versions of SATR: one without super points (denoted as \textit{SATR}) and one with super points (denoted as \textit{SATR+SP}).
    Since SATR does not perform instance segmentation, our comparison with it is focused on semantic segmentation.
    \item \textbf{SAM3D}~\cite{yang2023sam3d}  is another zero-shot 3D segmentation method using a 2D segmentation model, not GLIP~\cite{li2022glip}, but SAM~\cite{kirillov2023segment}. Since it only performs instance segmentation but does not label each segment, we compare our method with it only for instance segmentation, using the part-agnostic mAP metric.
\end{itemize}

\paragraph{\textbf{Ablation Study.}}
We also conduct an ablation study, comparing our method with two major components being ablated: the weight prediction (Sec.~\ref{sec:weight_prediction}) and the SAM mask integration (Sec.~\ref{sec:sam_mask}).
In the \textbf{supplementary material}, we also compare our method with cases using cross-entropy loss instead of mRIoU loss (Sec.~\ref{sec:mriou}) and using GLIP bounding box confidence score instead of our learned weights in the weighted voting (Sec.~\ref{sec:weight_prediction}). Note that if both weight prediction and SAM mask integration are excluded, our \projname{} is identical to PartSLIP~\cite{liu2023partslip}.

\subsection{Semantic Segmentation Results}
Tab.~\ref{tab:semantic_quantitative} and Fig.~\ref{fig:semantic_qualitative} present the quantitative and qualitative comparisons of part semantic segmentation results, respectively. Please refer to the supplementary material for the complete table encompassing all 45 categories. In comparison to PartSLIP~\cite{liu2023partslip}, a SotA few-shot 3D segmentation method, we achieve a significant \textbf{7.0\%p mIoU} improvement over the entire set of 45 categories. Furthermore, for each specific category, we consistently demonstrate improvement, surpassing by more than 15\%p in some categories (e.g., Remote). Qualitatively, the results also highlight the effectiveness of our task adaptation and the SAM mask integration in segmenting parts more precisely, even for small (e.g., Camera, Chair) and thin (e.g., Clock) parts of objects.

SATR exhibits inferior performance, even when leveraging the finetuned GLIP model and super points (see SATR+SP results) due to the label-propagation-based 2D bounding box integration, which often fails to provide clear boundaries in 3D segmentation, as illustrated in Fig.~\ref{fig:semantic_qualitative}.

The ablation study results demonstrate the influence of each major component in our framework. The quantitative findings indicate that the weight prediction network plays a crucial role in substantial improvement, leading to a 3.1\%p decrease in average mIoU when it is ablated. The SAM mask integration also contributes significantly, resulting in a 2.9\%p average mIoU decrease when ablated.

\begin{figure*}[t!]
    \centering
    \scriptsize{
        \renewcommand{\arraystretch}{0.0}
        \setlength{\tabcolsep}{0.0em}
        \setlength{\fboxrule}{0.0pt}
        \setlength{\fboxsep}{0pt}
        
        \begin{tabularx}{\textwidth}{>{\centering\arraybackslash}m{0.13\textwidth}>{\centering\arraybackslash}m{0.87\textwidth}}\\
         & \framebox{\includegraphics[width=0.87\textwidth]{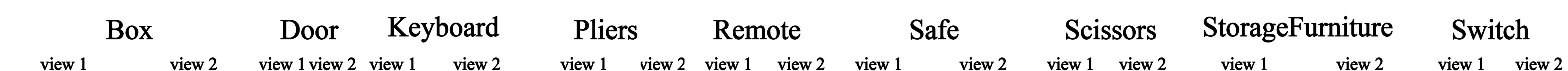}} \\
        {\tiny\makecell[c]{Input}} & \framebox{\includegraphics[width=0.87\textwidth]{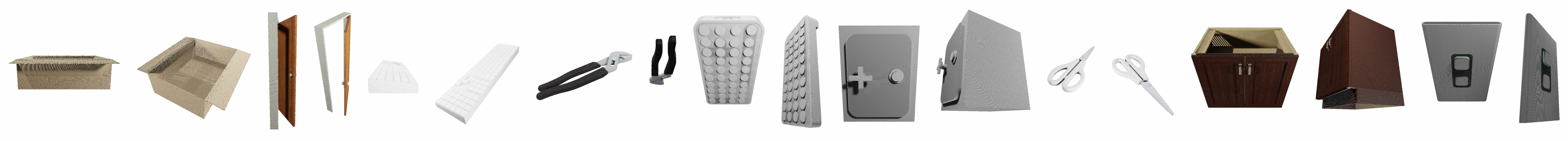}} \\
        {\tiny\makecell[c]{GT}} & \framebox{\includegraphics[width=0.87\textwidth]{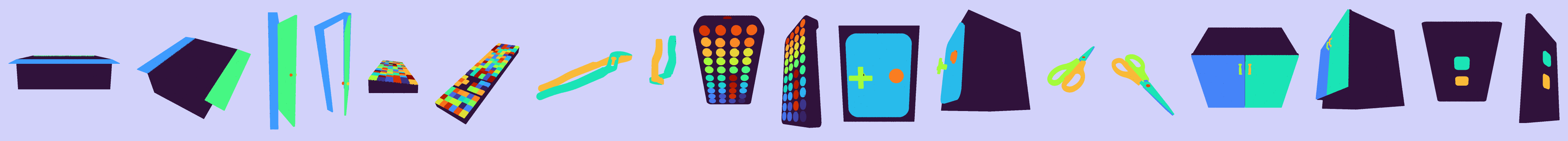}} \\
        {\tiny\makecell[c]{SAM3D~\cite{yang2023sam3d}}} & \framebox{\includegraphics[width=0.87\textwidth]{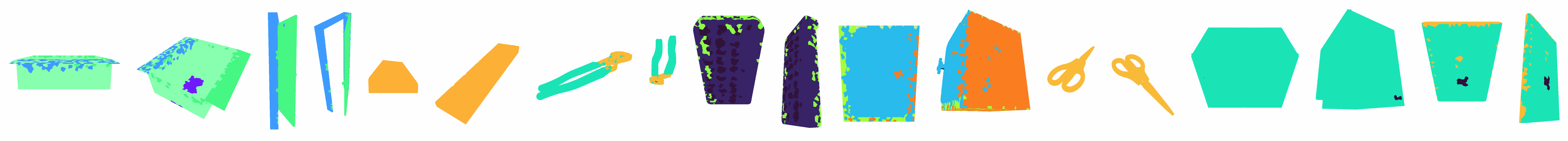}} \\
        {\tiny\makecell[c]{PartSLIP~\cite{liu2023partslip}}} & \framebox{\includegraphics[width=0.87\textwidth]{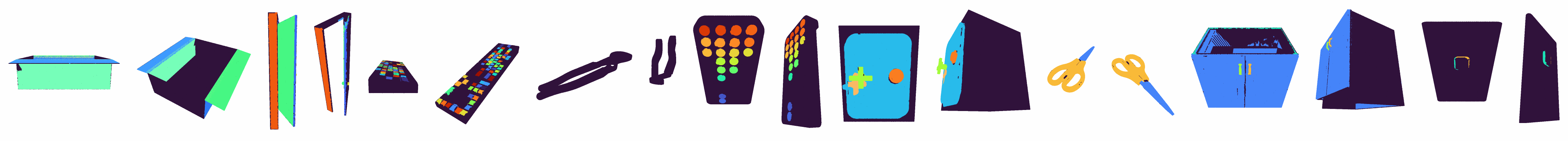}} \\ 
        {\tiny\makecell[c]{\projname{} \\(Ours)}} & \framebox{\includegraphics[width=0.87\textwidth]{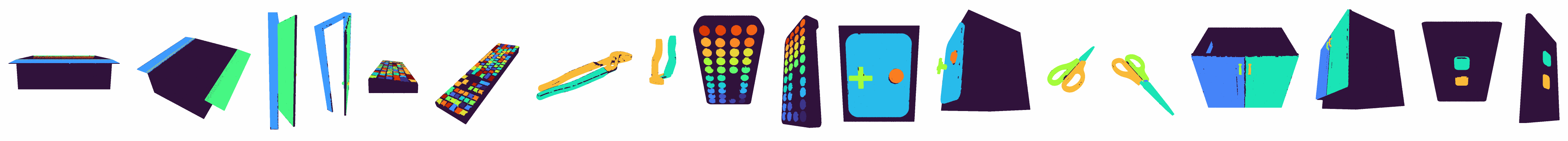}} \\
        \end{tabularx}
    }
    \caption{Qualitative comparison of instance segmentation results shows that our \projname{} successfully segments tiny 3D parts, such as the keys of keyboards and buttons of remote controls, with clear segment boundaries.}
    \label{fig:instance_qualitative}
\end{figure*}

\begin{table*}[t!]
    \caption{Part-aware mAPs(\%) of the instance segmentation results on the PartNet-Mobility dataset. Ours achieves \textbf{4.0\%p} improvement in mean part-aware mAPs compared to PartSLIP~\cite{liu2023partslip}, with consistent enhancements across all categories. Please refer to the supplementary material for the complete table with results for all 45 categories.}
    \centering
    {\scriptsize
    \setlength{\tabcolsep}{0.1em}
    \begin{tabularx}{\linewidth}{>{\centering\arraybackslash}m{0.2\textwidth}|>{\centering\arraybackslash}m{0.08\textwidth}|>{\centering\arraybackslash}m{0.09\textwidth}>{\centering\arraybackslash}m{0.07\textwidth}>{\centering\arraybackslash}m{0.07\textwidth}>{\centering\arraybackslash}m{0.07\textwidth}>{\centering\arraybackslash}m{0.07\textwidth}>{\centering\arraybackslash}m{0.07\textwidth}>{\centering\arraybackslash}m{0.07\textwidth}>{\centering\arraybackslash}m{0.07\textwidth}>{\centering\arraybackslash}m{0.07\textwidth}}\toprule
Method&mAP&{\tiny\makecell[b]{Storage\\Furniture}}&{\tiny Table}&{\tiny Chair}&{\tiny Switch}&{\tiny Toilet}&{\tiny Laptop}&{\tiny USB}&{\tiny Remote}&{\tiny Scissors}\\
\midrule
PartSLIP~\cite{liu2023partslip}&41.6&29.1&32.6&82.2&21.2&36.2&17.8&20.9&19.9&23.6\\
\midrule 
\multicolumn{11}{c}{Ablation Study} \\
\midrule
\textit{w/o Weight Pred.}&44.7&33.8&30.8&\textbf{82.5}&\textbf{22.2}&\textbf{40.8}&24.4&17.4&20.3&25.5\\
\textit{w/o SAM Integ.}&44.2&29.4&32.3&82.2&19.7&36.0&23.6&24.8&29.6&23.1\\
\midrule
\projname{}(Ours)&\textbf{45.6}&\textbf{35.5}&\textbf{33.2}&\textbf{82.5}&22.1&40.6&\textbf{28.9}&\textbf{26.5}&\textbf{33.7}&\textbf{26.5}\\
    \bottomrule
    \end{tabularx}
    \label{tab:part_aware_instance_quantitative}
    }
\end{table*}

\begin{table*}[ht!]
    \caption{Part-agnostic mAPs(\%) of the instance segmentation results on the PartNet-Mobility dataset. Ours demonstrates \textbf{5.2\%p} improvement in mean part-agnostic mAPs compared to PartSLIP~\cite{liu2023partslip}, surpassing SAM3D with a mean mAP that is more than three times larger. Please refer to the supplementary material for the complete table with results for all 45 categories. (*SAM3D is a zero-shot method.)}
    \centering
    {\scriptsize
    \setlength{\tabcolsep}{0.1em}
    \begin{tabularx}{\linewidth}{>{\centering\arraybackslash}m{0.2\textwidth}|>{\centering\arraybackslash}m{0.08\textwidth}|>{\centering\arraybackslash}m{0.09\textwidth}>{\centering\arraybackslash}m{0.07\textwidth}>{\centering\arraybackslash}m{0.07\textwidth}>{\centering\arraybackslash}m{0.07\textwidth}>{\centering\arraybackslash}m{0.07\textwidth}>{\centering\arraybackslash}m{0.07\textwidth}>{\centering\arraybackslash}m{0.07\textwidth}>{\centering\arraybackslash}m{0.07\textwidth}>{\centering\arraybackslash}m{0.07\textwidth}}\toprule
Method&mAP&{\tiny\makecell[b]{Storage\\Furniture}}&{\tiny Table}&{\tiny Chair}&{\tiny Switch}&{\tiny Toilet}&{\tiny Laptop}&{\tiny USB}&{\tiny Remote}&{\tiny Scissors}\\
\midrule
SAM3D*~\cite{yang2023sam3d}&12.1&1.2&10.6&5.5&5.4&3.7&1.5&22.6&1.0&7.2\\
PartSLIP~\cite{liu2023partslip}&38.9&29.7&28.7&80.7&21.2&35.4&19.5&20.5&19.9&27.0\\
\midrule 
\multicolumn{11}{c}{Ablation Study} \\
\midrule
\textit{w/o Weight Pred.}&42.6&35.9&\textbf{29.2}&81.1&22.2&40.0&26.5&15.2&20.3&\textbf{29.4}\\
\textit{w/o SAM Integ.}&42.6&34.8&27.6&81.2&19.7&36.6&27.4&\textbf{27.3}&29.6&24.9\\
\midrule
\projname{} (Ours)&\textbf{44.1}&\textbf{41.7}&28.2&\textbf{83.3}&\textbf{22.4}&\textbf{41.0}&\textbf{34.1}&23.5&\textbf{33.7}&26.2\\

    \bottomrule
    \end{tabularx}
    \label{tab:part_agnostic_instance_quantitative}
    }
\end{table*}

\subsection{Instance Segmentation Results}

Following the instance segmentation idea of PartSLIP, we also perform instance segmentation using our method and compare the quantitative results with those of other methods in Tab.~\ref{tab:part_aware_instance_quantitative} and Tab.~\ref{tab:part_agnostic_instance_quantitative}, reporting part-aware mAPs and part-agnostic mAPs, respectively.
Fig.~\ref{fig:instance_qualitative} also displays qualitative comparisons across different methods. Please refer to the supplementary material for the complete table of all 45 categories. Similar to semantic segmentation, ours achieves \textbf{4.0\%p} and \textbf{5.2\%p} improvements for both average part-aware mAPs and part-agnostic mAPs, respectively. Both the weight prediction network and SAM mask integration also exhibit meaningful improvements in performance, as their influence is depicted in the ablation study results. Compared to SAM3D~\cite{yang2023sam3d}, whose part-agnostic mAPs are shown in Tab.~\ref{tab:part_agnostic_instance_quantitative}, our method achieves much better performance with more than three times greater mAP. Qualitative results also illustrate the outstanding performance of our method, accurately identifying all the tiny instances of parts, such as keys on keyboards, buttons on remotes, and clear boundaries of these instances.

 \begin{figure}[t!]
    \centering
    \includegraphics[width=\textwidth]{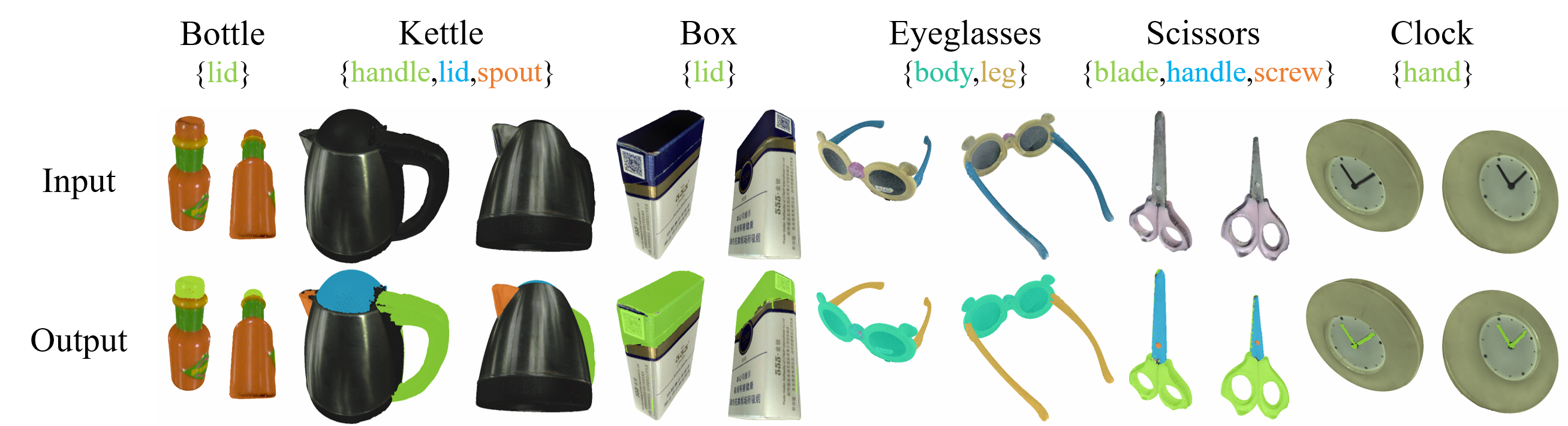}
    \caption{Qualitative comparison of semantic segmentation results on OmniObect3D~\cite{wu2023omniobject3d} dataset, a high quality real scanned 3D objects dataset. Our \projname{} predicts precise boundaries, even in the case that the object has an appearance significantly different from training data such as box category.}
    \label{fig:omniobject3d}
\end{figure}
\subsection{Part Segmentation with Real Scanned Dataset}
Fig.~\ref{fig:omniobject3d} displays additional results on the OmniObject3D~\cite{wu2023omniobject3d} dataset, a high-quality dataset featuring real-scanned 3D objects. We present qualitative results only, as ground truth segmentation for OmniObject3D~\cite{wu2023omniobject3d} is unavailable. Note that our \projname{} accurately identifies parts in real scans, demonstrating its effectiveness beyond synthetic data. Particularly impressive is its performance in the box category depicted in Fig.~\ref{fig:omniobject3d}, where despite significant discrepancies in appearance compared to the training data, \projname{} achieves remarkably accurate predictions.

\section{Conclusion and Future Work}
We presented \projname{}, a task adaptation method for lifting 2D segmentation to 3D. Instead of directly finetuning a 2D segmentation network, our method learns a small neural network predicting weights for each 2D bounding box with an objective function for 3D segmentation, and then performs 3D segmentation via weighted merging of the boxes in 3D. We further improve the performance of 3D segmentation by integrating SAM foreground masks for each bounding box. We achieve the SotA results in few-shot 3D part segmentation, demonstrating significant improvements in both semantic and instance segmentations.

As \projname{} still depends on 2D representations for 3D predictions, its understanding of 3D geometry is relatively limited, and thus it cannot account for occluded or interior points, which poses a limitation. In future work, we plan to further investigate combining our approach with the direct utilization of 3D representations, such as leveraging 3D features, for more accurate predictions.

\section*{\textbf{Acknowledgements}}
We sincerely thank Minghua Liu for providing the code of PartSLIP and answering questions. This work was supported by NRF grant (RS-2023-00209723), IITP grants (2022-0-00594, RS-2023-00227592, RS-2024-00399817), and Alchymist Project Program (RS-2024-00423625) funded by the Korean government (MSIT and MOTIE), and grants from the DRB-KAIST SketchTheFuture Research Center, NAVER-intel, Adobe Research, Hyundai NGV, KT, and Samsung Electronics.

\bibliographystyle{splncs04}
\bibliography{main}

\clearpage
\newpage
\renewcommand{\thesection}{S.\arabic{section}}
\renewcommand{\thetable}{S\arabic{table}}
\renewcommand{\thefigure}{S\arabic{figure}}
\setcounter{section}{0}
\setcounter{table}{0}
\setcounter{figure}{0}
\title{PartSTAD: 2D-to-3D Part Segmentation Task Adaptation} 
\author{\textbf{Supplementary Material}}
\authorrunning{H. Kim and M. Sung}
\maketitle

\makeatletter
\newcommand{\manuallabel}[2]{\def\@currentlabel{#2}\label{#1}}
\makeatother

\manuallabel{subsec:mriou_loss}{4.1}
\manuallabel{subsec:weight_pred_network}{4.2}
\manuallabel{subsec:result_real_scanned}{5.4}

\manuallabel{eq:voting_formula}{5}
\manuallabel{eq:modified_relu}{7}

\manuallabel{fig:pipeline}{2}
\manuallabel{fig:omniobject3d_supp}{5}
\manuallabel{tbl:quantitative}{1}
\manuallabel{tbl:quantitative2}{2}

\newcommand{\refofpaper}[1]{of the main paper}
\newcommand{\refinpaper}[1]{in the main paper}

\noindent
In this supplementary material, we present additional implementation details (Sec.~\ref{sec: additional implementation detail}), additional results on the ablation studies (Sec.~\ref{sec: varying parameters}, Sec.~\ref{sec: viewpoints}, Sec.~\ref{sec: loss ablation}, Sec.~\ref{sec: vanilla GLIP vs fine-tuned GLIP}, and Sec.~\ref{sec: loss confidence}), how bounding box weights change compared to GLIP confidence scores (Sec.~\ref{sec:bbox_weight_change}), the reason for not replacing GLIP with SAM (Sec.~\ref{sec: reason for sam}), more analysis on results (Sec.~\ref{sec: more analysis on results}), outcomes with scanned data (Sec.~\ref{sec: Results on Real-world Data}), and comprehensive results covering all PartNet-Mobility categories (Sec.~\ref{sec: Whole Quantitative Result}, Sec.~\ref{sec: Additional Qualitative Results Semantic}, and Sec.~\ref{sec: Additional Qualitative Results Instance}).

 \begin{figure}[h!]
    \centering
    \includegraphics[width=0.9\textwidth]{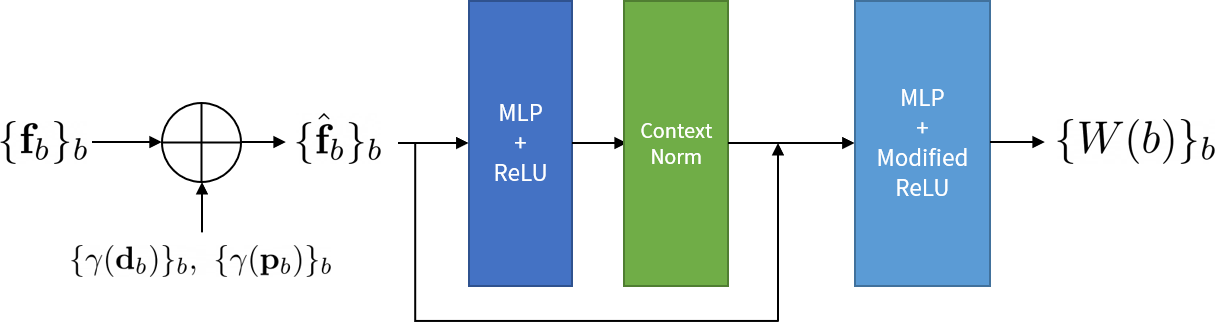}
    \caption{Architecture of the weight prediction network used in \projname{}. $\mathbf{f}_b$, $\mathbf{d}_b$, and $\mathbf{p}_b$ represent the bounding box feature, view direction, and position in the 2D image of the bounding box $b$, respectively. $\gamma$ denotes the positional encoding function. This network takes all bounding box features of a single object as input and outputs bounding box weights.}
    \label{fig:network_architecture}
\end{figure}
\section{Additional Implementation Details}
\label{sec: additional implementation detail}
\paragraph{\textbf{Network Architecture.}}
Fig.~\ref{fig:network_architecture} shows the detailed network architecture of the weight prediction network of \projname{}. As mentioned in Sec.~\ref{subsec:weight_pred_network}~\refofpaper{}, the network consists of small shared two-layer MLP with the context normalization~\cite{yi2018learningCNe} layer between them to embed context information. This light architecture design is inspired by LoRA~\cite{hu2021lora} and PartSLIP~\cite{liu2023partslip} which add small learnable parameters while keeping the original pretrained model parameters. 

We add positional encoded vectors to each bounding box feature to embed the positional information. For given bounding box feature $\mathbf{f}_b$, we do not directly feed $\mathbf{f}_b$ to the network but feed $\hat{\mathbf{f}}_b$ which concatenates $\mathbf{f}_b$, positional encoding of view direction $\mathbf{d}_b$, and positional encoding of 2D bounding box position $\mathbf{p}_b$. Positional encoding $\gamma$ is defined as below:
\begin{align}
    \gamma(x_1, x_2, ..., x_n) = \bigoplus_{i=1}^{n}(x_i, \sin(2^0\pi x_i), \cos(2^0\pi x_i), ... ,\sin(2^{L-1}\pi x_i), \cos(2^{L-1}\pi x_i)),
\end{align}
where $\oplus$ indicates the concatenation operation and $L$ is set to 10 in our experiments. Thus the input $\hat{\mathbf{f}}_b$ of the weight prediction network can be written as below:
\begin{align}
    \hat{\mathbf{f}}_b = \mathbf{f}_b \oplus \gamma(\mathbf{d}_b) \oplus \gamma(\mathbf{p}_b).
\end{align}

We initialize all network parameters $\theta_i \sim \mathcal{N}(0, \epsilon)$, where $\epsilon$ is a very small number ($\epsilon$ is set to 0.0001 in our experiments) so that the initial MLP output becomes 0. Since the last layer is modified ReLU layer $\phi$ (Eq.~\ref{eq:modified_relu}~\refofpaper{}), the initial weight is set to $\tau$. This is inspired by zero convolution of ControlNet~\cite{zhang2023controlnet}, and this makes the training more stable by preventing drastic weight changes. 

Instead of using an attention-based network to consider the relations between bounding boxes, we opt to add context normalization~\cite{yi2018learningCNe} between two MLP networks in the weight prediction network. This allowed us to keep the network lightweight while still considering the relations between bounding boxes.
\paragraph{\textbf{Design of Modified ReLU.}}
The modified ReLU function (Eq.~\ref{eq:modified_relu}~\refofpaper{}) is designed to set the initial value of the bbox weight, which is the output of the weight prediction network, to a value $\tau > 0$. 

\paragraph{\textbf{Training Details.}}
We generate 2D images of size $800\times800$ from 10 fixed viewpoints for each object that is normalized to fit in a unit sphere using the Pytorch3D point cloud renderer with the fixed camera distance 2.2 following the same procedure as described in PartSLIP~\cite{liu2023partslip}. After rendering, we obtain bounding boxes for each image using the GLIP~\cite{li2022glip}. Subsequently, all bounding box features corresponding to bounding boxes from a single object are simultaneously fed into the weight prediction network to calculate the weights. Training is conducted using a single RTX 3090 GPU.

\section{Results with Varying Parameters}
\label{sec: varying parameters}
\begin{table}[h!]
\caption{Ablation study on the number of views.}
\centering
\scriptsize{
\setlength{\tabcolsep}{0.2em}
\renewcommand{\arraystretch}{1.0}
\begin{tabularx}{0.8\linewidth}{>{\centering\arraybackslash}m{0.128\linewidth}|>{\centering\arraybackslash}m{0.16\linewidth}>{\centering\arraybackslash}m{0.16\linewidth}>{\centering\arraybackslash}m{0.16\linewidth}>{\centering\arraybackslash}m{0.16\linewidth}}
  \toprule

\# of Views & 5 & 10 & 15 & 20 \\
\midrule
mIoU & 59.4 & 65.0 & 66.3 & \textbf{67.2} \\
  \bottomrule
\end{tabularx}
}
\label{tab:supp_view}
\end{table}

\begin{table}[h!]
\caption{Ablation study on the number of training data. The experiment is conducted on the StorageFurniture category as it only has more than 128 shapes (346 in total).}
\centering
\scriptsize{
\setlength{\tabcolsep}{0.2em}
\renewcommand{\arraystretch}{1.0}
\begin{tabularx}{0.8\linewidth}{>{\centering\arraybackslash}m{0.216\linewidth}|>{\centering\arraybackslash}m{0.104\linewidth}>{\centering\arraybackslash}m{0.104\linewidth}>{\centering\arraybackslash}m{0.104\linewidth}>{\centering\arraybackslash}m{0.104\linewidth}>{\centering\arraybackslash}m{0.104\linewidth}}
  \toprule

\# of Training Data & 8 & 16 & 32 & 64 & 128 \\
\midrule
mIoU & 57.0 & 56.7 & 58.0 & 57.8 & \textbf{60.1} \\
  \bottomrule
\end{tabularx}
}
\label{tab:supp_instance}
\end{table}

\begin{table}[h!]
\caption{Ablation study on the hyperparameter $\tau$. The mIoU is measured for five object categories: Chair, Table, StorageFurniture, Faucet, and TrashCan, which are the five categories with the most test data.}
\centering
\scriptsize{
\setlength{\tabcolsep}{0.2em}
\renewcommand{\arraystretch}{1.0}
\begin{tabularx}{0.9\linewidth}{>{\centering\arraybackslash}m{0.316\linewidth}|>{\centering\arraybackslash}m{0.104\linewidth}>{\centering\arraybackslash}m{0.104\linewidth}>{\centering\arraybackslash}m{0.104\linewidth}>{\centering\arraybackslash}m{0.104\linewidth}>{\centering\arraybackslash}m{0.104\linewidth}}
  \toprule

Initial Weight ($\tau$) & 1 & 5 & 10 & 15 & 20 \\
\midrule
mIoU (5 categories) & 54.8 & 55.6 & \textbf{55.8} & 55.6 & 55.6 \\
  \bottomrule
\end{tabularx}
}
\label{tab:tau}
\end{table}

\paragraph{\textbf{Number of Views.}}
Tab.~\ref{tab:supp_view} presents the results of the ablation study on the number of views. It is observed that as the number of views increases, mIoU also increases, with the most significant difference observed when the view changes from 5 to 10. This highlights that having too few samples of bounding boxes used in training can lead to suboptimal results.

\paragraph{\textbf{Number of Training Data.}}
Tab.~\ref{tab:supp_instance} presents the results of the ablation study based on the number of training data. Only the StorageFurniture category has more than 128 data, so the experiment is conducted only for this category. There is a tendency for mIoU to increase as the number of training data increases, but the difference is not significant. This demonstrates that even using only 8 data points can yield sufficiently good results.

\paragraph{\textbf{Hyperparameter $\tau$.}} 
Tab.~\ref{tab:tau} shows the ablation results for the hyperparameter $\tau$. The results are best when $\tau$ is 10, but they also demonstrate that the $\tau$ value does not significantly impact the results when it is greater than zero.

\section{\textbf{Random Viewpoints vs. Fixed Viewpoints}}
\label{sec: viewpoints}
\begin{table}[h!]
\caption{Comparison between cases of rendering with random viewpoints and fixed viewpoints.}
\centering
\scriptsize{
\setlength{\tabcolsep}{0.2em}
\renewcommand{\arraystretch}{1.0}
\begin{tabularx}{0.9\linewidth}{>{\centering\arraybackslash}m{0.266\linewidth}|>{\centering\arraybackslash}m{0.114\linewidth}>{\centering\arraybackslash}m{0.114\linewidth}>{\centering\arraybackslash}m{0.114\linewidth}>{\centering\arraybackslash}m{0.114\linewidth}>{\centering\arraybackslash}m{0.114\linewidth}}
  \toprule

Category & Chair & Kettle & Lamp & Suitcase & Scissors \\
\midrule
Random & \textbf{85.5} & 82.5 & 66.8 & \textbf{71.3} & 64.8 \\
Fixed (Ours) & 85.3 & \textbf{84.2} & \textbf{68.4} & 68.3 & \textbf{68.5} \\
  \bottomrule
\end{tabularx}
}
\label{tbl:supp_randomview}
\end{table}

Tab.~\ref{tbl:supp_randomview} shows the results with random viewpoints and shows that the outcomes are not sensitive to the choice of viewpoints.

\section{Cross-Entropy Loss vs. mRIoU Loss}
\label{sec: loss ablation}
\begin{table}[h!]
\caption{Comparison with the cases of using cross-entropy loss and mRIoU loss (Ours).}
\centering
\footnotesize{
\setlength{\tabcolsep}{0.2em}
\renewcommand{\arraystretch}{1.0}
\begin{tabularx}{\linewidth}{>{\centering\arraybackslash}m{0.5\linewidth}|>{\centering\arraybackslash}m{0.5\linewidth}}
  \toprule

Method & mIoU (\%) \\
\midrule
{\scriptsize PartSLIP~\cite{liu2023partslip} + \textit{SAM Mask Integration} (Baseline)} & 61.9\\
\midrule
{\scriptsize\projname{} + Cross-Entropy - mRIoU} & 63.5\\ 
{\scriptsize\projname{} + Cross-Entropy} & 64.5\\ 
{\scriptsize\projname{} (Ours)} & \textbf{65.0}\\ 
  \bottomrule
\end{tabularx}
}
\label{tbl:supp_ablation_loss}
\end{table}

As mentioned in Sec.~\ref{subsec:mriou_loss}~\refofpaper{}, the use of mRIoU loss is crucial for achieving significant improvement in our task adaptation. To demonstrate the effectiveness of our mRIoU loss, we conduct an experiment comparing it with the alternative, cross-entropy loss.

Tab.~\ref{tbl:supp_ablation_loss} shows the ablation results for different loss types. Baseline at the $1$st row in Tab.~\ref{tbl:supp_ablation_loss} represents the result which only applies \textit{SAM}~\cite{kirillov2023segment} \textit{mask integration} to PartSLIP~\cite{liu2023partslip} (same as our method \textit{without weight prediction}). When using the commonly used cross-entropy loss for segmentation tasks, the mIoU decreases by 1.5\%p compared to using the mRIoU loss. Even when both losses are used together, the mIoU decreases by 0.5\%p. This indicates that the mRIoU loss is more effective for 3D segmentation task adaptation, and it shows its highest effectiveness when used alone.

\section{Vanilla GLIP vs. Finetuned GLIP}
\label{sec: vanilla GLIP vs fine-tuned GLIP}
\begin{figure}[h!]
    \centering
    \includegraphics[width=0.95\textwidth]{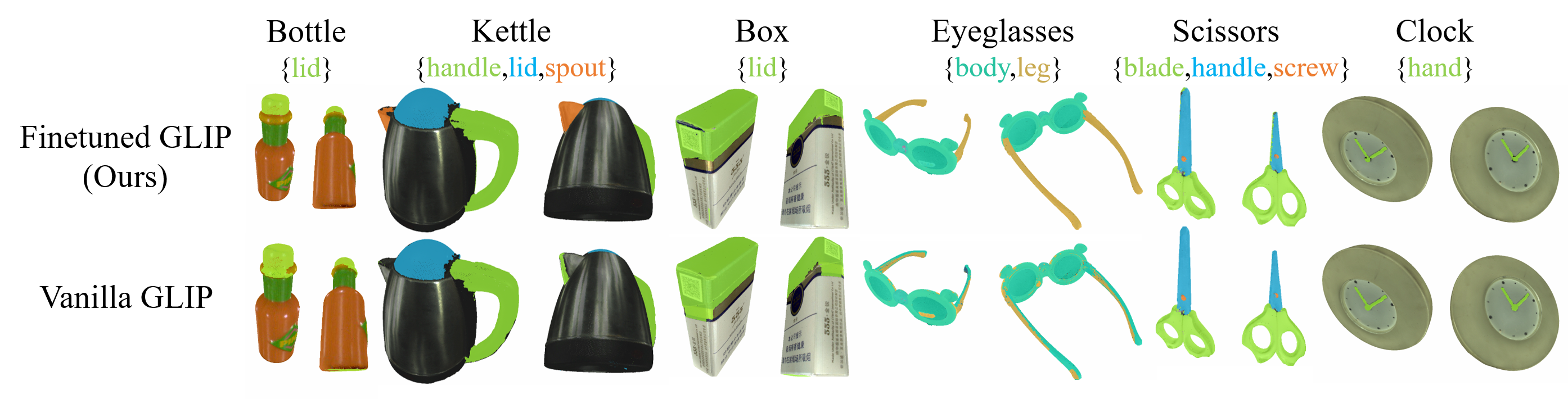}
    \caption{Qualitative comparison between vanilla GLIP and finetuned GLIP on OmniObject3D~\cite{wu2023omniobject3d}, a real-scanned dataset.}
    \label{fig:supp_omniobject3d_vanilla_vs_finetuned}
\end{figure}

\begin{table}[h!]
\caption{Quantitative comparison between vanilla GLIP and finetuned GLIP on PartNet-Mobility~\cite{Xiang_2020_SAPIEN} dataset.}
\centering
\scriptsize{
\setlength{\tabcolsep}{0.2em}
\renewcommand{\arraystretch}{1.0}
\begin{tabularx}{0.8\linewidth}{>{\centering\arraybackslash}m{0.24\linewidth}|>{\centering\arraybackslash}m{0.28\linewidth}>{\centering\arraybackslash}m{0.28\linewidth}}
  \toprule

Method & Vanilla GLIP & Finetuned GLIP (Ours) \\
\midrule
PartSLIP~\cite{liu2023partslip} & 27.2 & 58.0 \\
\projname{} (Ours) & 48.9 \textcolor{color_green}{(+ 21.7)} & \textbf{65.0 \textcolor{color_green}{(+ 7.0)}} \\
  \bottomrule
\end{tabularx}
}
\label{tab:supp_glip}
\end{table}

Tab.~\ref{tab:supp_glip} compares results using vanilla GLIP and finetuned GLIP. Vanilla GLIP yields significantly worse results, emphasizing the substantial impact of bounding box prediction on the final outcome. At the same time, when we use the vanilla GLIP, our weight prediction network significantly improves the mIoU from 27.2 to 48.9. This indicates that our weight prediction network is more effective when the 2D prediction is inaccurate.

In our experiments with scanned objects (Sec.~\ref{subsec:result_real_scanned}~\refofpaper{}), we used a finetuned GLIP instead of the vanilla GLIP, as it exhibited better performance in detecting the parts, even for real images, due to its finetuning for the specific parts. Fig.~\ref{fig:supp_omniobject3d_vanilla_vs_finetuned} illustrates the superior performance of the finetuned GLIP compared to the vanilla GLIP for the OmniObject3D dataset.

\section{GLIP Confidence Score vs. Mask Weight}
\label{sec: loss confidence}
\newcolumntype{Z}{>{\centering\arraybackslash}m{0.09\textwidth}}
\begin{figure*}[ht!]
    \centering
    \scriptsize{
        \renewcommand{\arraystretch}{0.0}
        \setlength{\tabcolsep}{0.0em}
        \setlength{\fboxrule}{0.0pt}
        \setlength{\fboxsep}{0pt}
        
        \begin{tabularx}{0.9\textwidth}{ZZZZZZZZZZ}\\
        \toprule
        \multicolumn{2}{c}{\multirow{2}{*}{Input}} & \multicolumn{2}{c}{\multirow{2}{*}{GT}} & \multicolumn{3}{c}{GLIP Confidence} & \multicolumn{3}{c}{Mask Weight (Ours)}\\[0.5em]
        \cmidrule(lr){5-7}\cmidrule(lr){8-10}
        
         &  &  &  & Top 1 (view 1) & \multicolumn{2}{c}{Seg. Result} & Top 1 (view 2) & \multicolumn{2}{c}{Seg. Result} \\
         \midrule
         view 1& view 2& view 1& view 2& & view 1& view 2& & view 1& view 2\\
        \multicolumn{10}{c}{\includegraphics[width=0.9\textwidth]{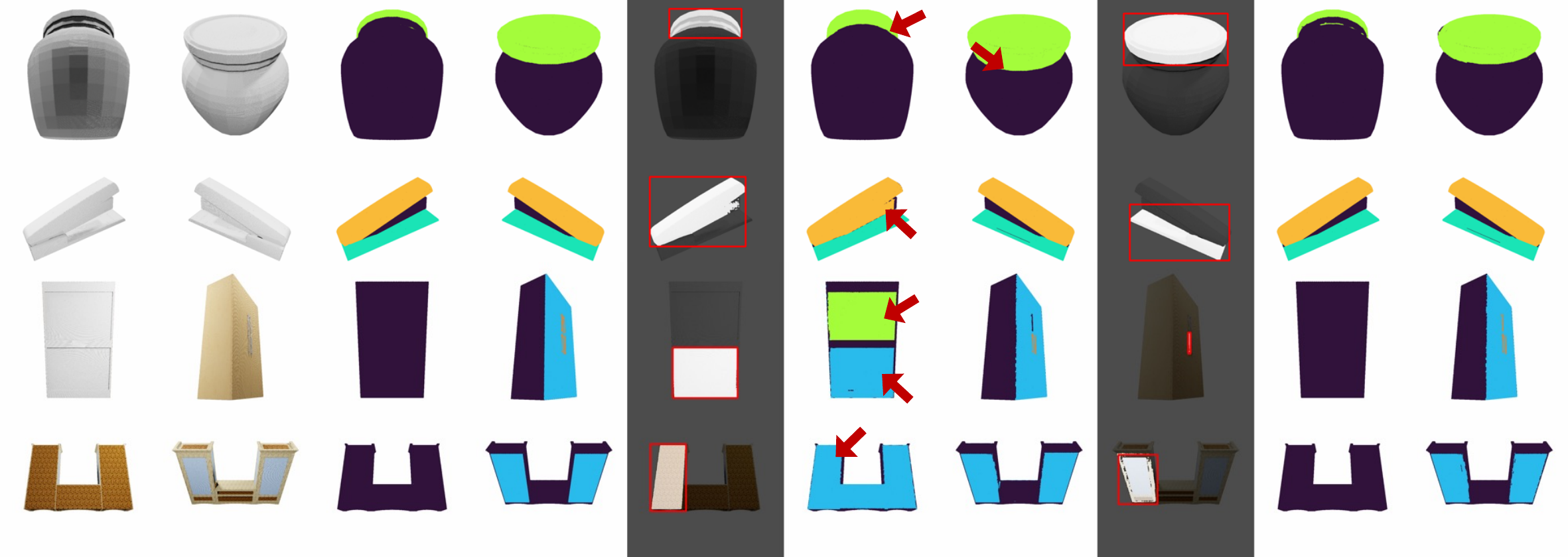}}
        \end{tabularx}
    }
    \caption{
Comparison of results using GLIP confidence scores and Mask Weights. The $5$th and the $8$th columns depict masks with the highest scores (weights), where red rectangles represent bounding boxes from GLIP, and the white regions are segmentation masks after integrating SAM. From the top row to the bottom, each corresponds to the Bottle, Stapler, and two StorageFurniture categories. When using GLIP confidence, the highest score mask for Bottle ($1$st row) and Stapler ($2$nd row) includes an incorrect region, leading to inaccurate segmentation (denoted as red arrows). In contrast, our method assigns the highest score to the correct mask, indicating that the incorrect mask has a lower score. Additionally, when using the GLIP confidence score, the highest score mask for the $3$rd and the $4$th rows each indicates a completely wrong part (the backside of StorageFurniture). However, our method assigns the highest score to the handle and the correct door part at the front side for the $3$rd row and the $4$th row, respectively.}
    \label{fig:ablation_glip_confidence_figure}
\end{figure*}

\begin{table}[h!]
\caption{Comparison with the cases of using GLIP confidence score as weight and predicted mask weight (Ours).}
\centering
\footnotesize{
\setlength{\tabcolsep}{0.2em}
\renewcommand{\arraystretch}{1.0}
\begin{tabularx}{\linewidth}{>{\centering\arraybackslash}m{0.5\linewidth}|>{\centering\arraybackslash}m{0.5\linewidth}}
  \toprule

Method & mIoU (\%) \\
\midrule
{\scriptsize PartSLIP~\cite{liu2023partslip} + \textit{SAM Mask Integration} (Baseline)} & 61.9\\
\midrule
{\scriptsize\projname{} + GLIP Conf.} & 53.3\\ 
{\scriptsize\projname{} + Normalized GLIP Conf.} & 62.3\\ 
{\scriptsize\projname{} (Ours)} & \textbf{65.0}\\ 
  \bottomrule
\end{tabularx}
}
\label{tbl:supp_ablation_confidence}
\end{table}

 It is worth noting that the GLIP~\cite{li2022glip} model also outputs a confidence score for each predicted bounding box. This implies that we can consider using the GLIP confidence scores as weights in the voting scheme ($W(b)$ in Eq.~\ref{eq:voting_formula}~\refofpaper{}). Thus, we compare the results when using GLIP confidence scores and the predicted mask weights from our method.
 
Tab.~\ref{tbl:supp_ablation_confidence} presents the comparison results between using GLIP confidence as weights and using weights predicted from the network. We compare those methods with the baseline ($1$st row in Tab.~\ref{tbl:supp_ablation_confidence}) which only applies \textit{SAM}~\cite{kirillov2023segment} \textit{mask integration} to PartSLIP~\cite{liu2023partslip} (same as our method \textit{without weight prediction}). As the confidence scores from GLIP range in $[0,1]$, utilizing them directly as weights causes an overall score reduction, resulting in segments are not generated. Consequently, the outcome is notably poor, with a 53.3 mIoU. To ensure a fair comparison, we normalize the weights to maintain the same sum as before. With these normalized weights, as presented in the second row of Tab.~\ref{tbl:supp_ablation_confidence}, the result becomes 62.3 mIoU, a slight increase of 0.4\%p compared to the baseline. However, this is still 2.7\%p lower than utilizing predicted mask weights from a network trained with 3D mRIoU loss. In conclusion, our method provides results more optimized for 3D segmentation than GLIP confidence scores, demonstrating the effectiveness of our method.

As shown in Fig~\ref{fig:ablation_glip_confidence_figure}, GLIP confidence scores occasionally assign the highest score to incorrect bounding boxes, leading to suboptimal segmentation results. In contrast, the weights predicted by our method consistently assign the highest weight to the correct regions. For instance, using the GLIP confidence score as the weight results in the highest score masks for Bottle (1st row) and Stapler (2nd row) including incorrect regions, leading to inaccuracies in segmentation (indicated by red arrows). In contrast, our method assigns the highest score to the correct mask, indicating that the incorrect mask has a lower score. Additionally, with the GLIP confidence score, the highest score masks for the $3$rd and the $4$th rows each indicate completely wrong parts (the backside of StorageFurniture). However, our method assigns the highest score to the handle and the correct door part at the front side for the $3$rd row and the $4$th row, respectively. Those results demonstrate that utilizing the mask weight predicted from the network trained with 3D mRIoU loss produces more accurate predictions compared to using the GLIP confidence score as the weight.

\section{\textbf{Learned Bounding Box Weights}}
\label{sec:bbox_weight_change}
 \begin{figure}[t!]
    \centering
\includegraphics[width=0.8\textwidth]{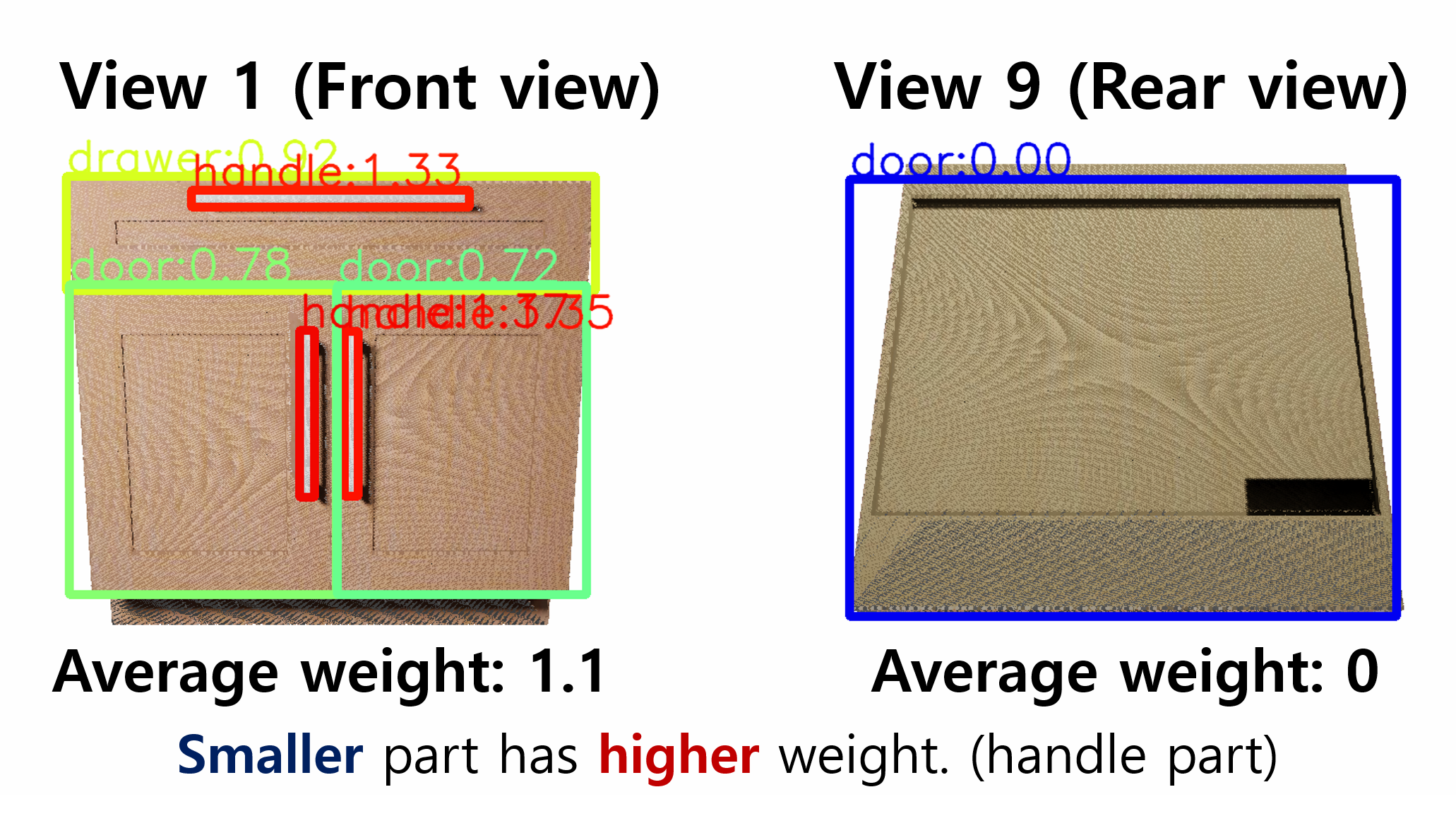}

    \caption{The smaller part (handle) has a higher weight compared to the bigger part (door), and the rear view (fewer GT parts) has a lower weight compared to the front view (more GT parts).}

    \label{fig:supp_learnedweight_example}
\end{figure}

As seen in the example of Fig.~\ref{fig:supp_learnedweight_example} for the storage furniture category, smaller parts like handles tend to have relatively higher weights compared to larger parts like doors. Additionally, in rear views where there are no ground truth parts, there is a tendency for the average weight to be lower compared to front views with many ground truth parts. This indicates that learned weight is influenced by both view direction and part labels, unlike the GLIP confidence score, which has a uniform average without distinct trends regarding view and parts.

\section{Reasons for Not Replacing GLIP~\cite{li2022glip} with SAM~\cite{kirillov2023segment}}
\label{sec: reason for sam}
SAM~\cite{kirillov2023segment} allows text prompts as inputs, which could enable direct replacement of GLIP with SAM. However, since the pretrained model supporting text prompts has not been released, we resort to an alternative approach. We serialize GLIP and SAM by using a bounding box predicted by GLIP as an input prompt for SAM.

Note that Grounded-SAM~\cite{ren2024grounded} and other recent 2D segmentation methods based on text prompts (e.g., SAM-HQ~\cite{sam_hq}) also involve the serialization of a bounding box prediction network (like GLIP) and SAM. Grounded-SAM specifically uses GroundingDINO~\cite{liu2023groundingdino} instead of GLIP. We believe that the selection of the bounding box prediction network should not impact our contributions.

\section{More Analysis on Results}
\label{sec: more analysis on results}
\begin{figure}
\begin{subfigure}{.5\textwidth}
  \centering
  \includegraphics[width=.6\linewidth]{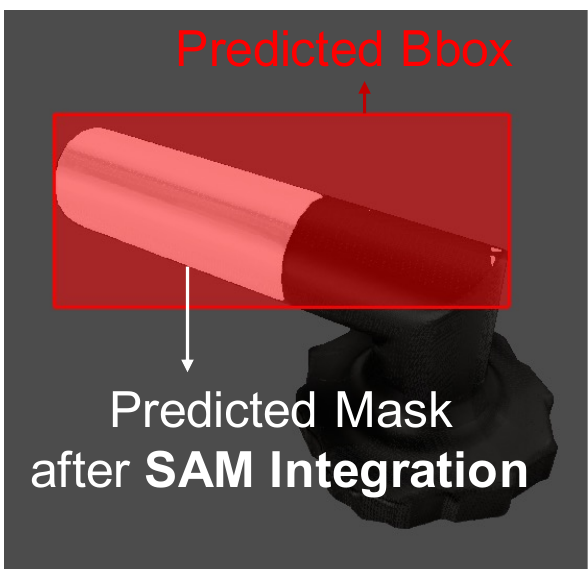}
  \caption{Failure case of SAM Integration.}
  \label{fig:sfig1}
\end{subfigure}%
\begin{subfigure}{.5\textwidth}
  \centering
  \includegraphics[width=.6\linewidth]{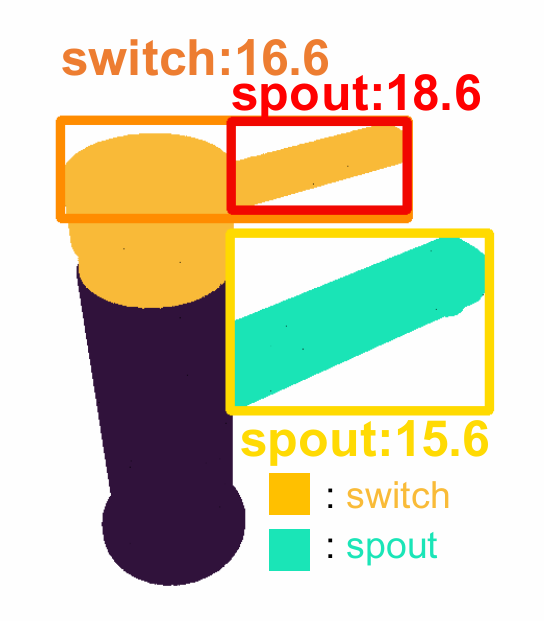}
  \caption{Failure case of Weight Prediction.}
  \label{fig:sfig2}
\end{subfigure}
\caption{Failure cases of \projname{}.}
\label{fig:failure}
\end{figure}
The qualitative and quantitative results in the main paper demonstrate that our \projname{} provides more specialized 2D predictions tailored to 3D segmentation compared to PartSLIP~\cite{liu2023partslip}. However, in some categories, there are cases where removing specific components from \projname{} leads to better results or even where PartSLIP outperforms \projname{} (e.g., Faucet category). Fig.~\ref{fig:failure} illustrates cases where each component performs worse predictions than the baseline.

When the initially given bounding box contains few wrong points but includes many correct points, there are cases where the new mask obtained through SAM does not include the previously contained correct points (Fig.~\ref{fig:sfig1}). In such cases, the performance may deteriorate when \textit{SAM mask integration} is applied. In the Faucet category, both the switch part and the spout part protrude prominently, resulting in the initial bounding box containing few irrelevant points. Therefore, it appears that the performance deteriorates when correct points are excluded rather than irrelevant points through SAM mask integration.

Secondly, in visually similar parts, weights might be inaccurately predicted. Fig.~\ref{fig:sfig2} illustrates the predicted bounding boxes and weights of the Faucet object, showing that the switch part is predicted as the spout with the highest bounding box weight. In such cases where parts are not visually distinguishable, the weight prediction may not be properly learned. Additionally, adding weight prediction in these cases may lead to a decrease in performance.

Note that such cases are rare and do not significantly impact the overall improvement, as shown in Tab.~\ref{tbl:quantitative} and Tab.~\ref{tbl:quantitative2}~\refinpaper{}.

Additionally, for some parts such as door and drawer sometimes have extremely low IoU. This is mainly caused by GLIP, as it fails to detect the parts due to a lack of data. For example, in the Table class, there is no training data that includes the door part.

\begin{figure*}[ht!]
    \centering
    \scriptsize{
        \renewcommand{\arraystretch}{0.0}
        \setlength{\tabcolsep}{0.0em}
        \setlength{\fboxrule}{0.0pt}
        \setlength{\fboxsep}{0pt}
        
        \begin{tabularx}{0.95\textwidth}{>{\centering\arraybackslash}m{0.15\textwidth}|>{\centering\arraybackslash}m{0.18\textwidth}>{\centering\arraybackslash}m{0.199\textwidth}>{\centering\arraybackslash}m{0.103\textwidth}>{\centering\arraybackslash}m{0.161\textwidth}>{\centering\arraybackslash}m{0.151\textwidth}}
        Category & Cart & Chair & Dispenser & Display & Faucet \\[1em]
         
        \makecell[c]{Text Prompt} & 
        {\scriptsize\makecell[c]{\textcolor{color_lightgreen}{wheel}}} & 
        {\scriptsize\makecell[c]{\textcolor{color_deepblue}{arm},\textcolor{color_cyan}{back},\textcolor{color_lightgreen}{leg},\\ \textcolor{color_yellow}{seat},\textcolor{color_orange}{wheel}}} & 
        {\scriptsize\makecell[c]{\textcolor{color_yellow}{head},\textcolor{color_lightgreen}{lid}}} & 
        {\scriptsize\makecell[c]{\textcolor{color_skyblue}{base},\textcolor{color_lightgreen}{screen},\\ \textcolor{color_orange}{support}}} & 
        {\scriptsize\makecell[c]{\textcolor{color_cyan}{spout},\textcolor{color_yellow}{switch}}} \\[1em]
        \midrule
        Input & \includegraphics[width=0.18\textwidth]{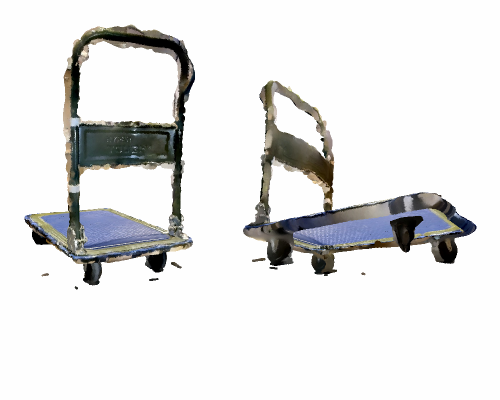} & \includegraphics[width=0.199\textwidth]{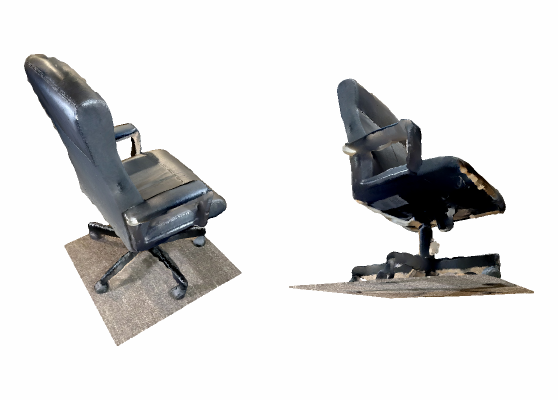} & \includegraphics[width=0.103\textwidth]{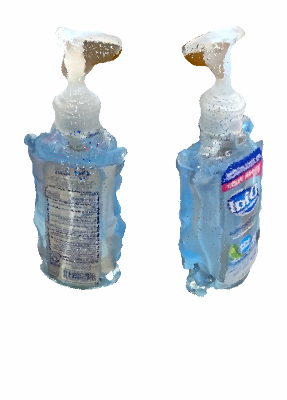} & \includegraphics[width=0.161\textwidth]{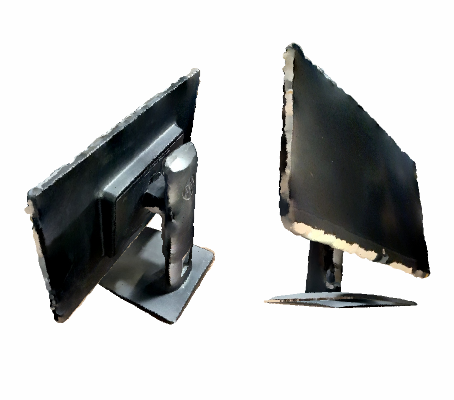} & \includegraphics[width=0.151\textwidth]{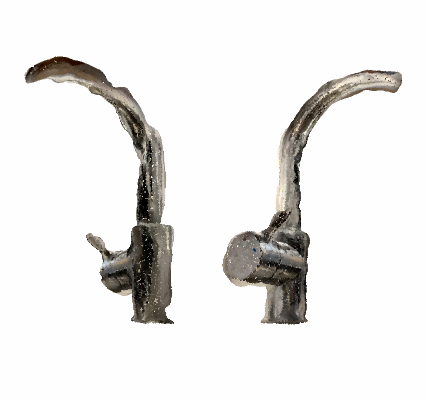} \\
        PartSLIP~\cite{liu2023partslip} & \includegraphics[width=0.18\textwidth]{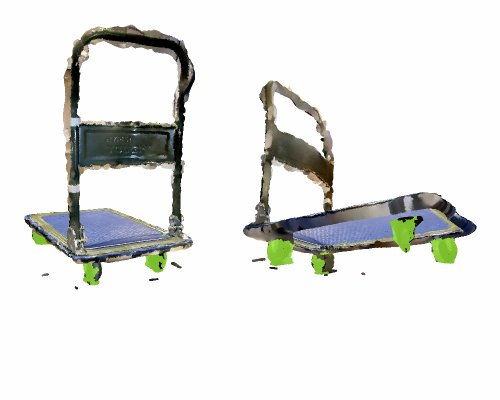} & \includegraphics[width=0.199\textwidth]{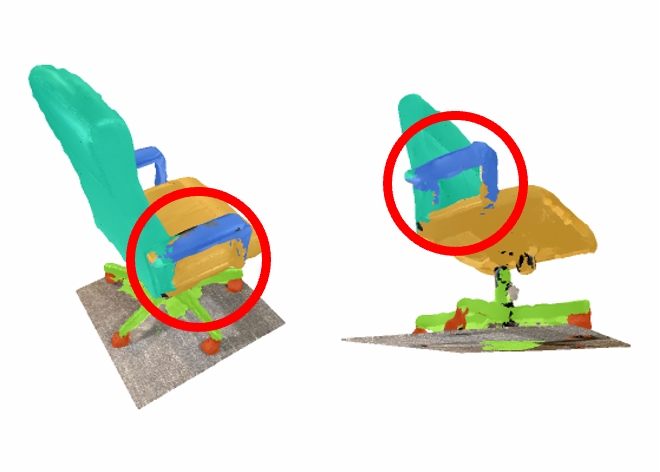} & \includegraphics[width=0.103\textwidth]{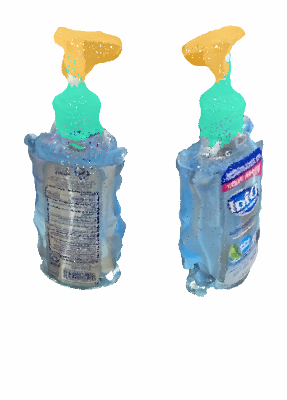} & \includegraphics[width=0.161\textwidth]{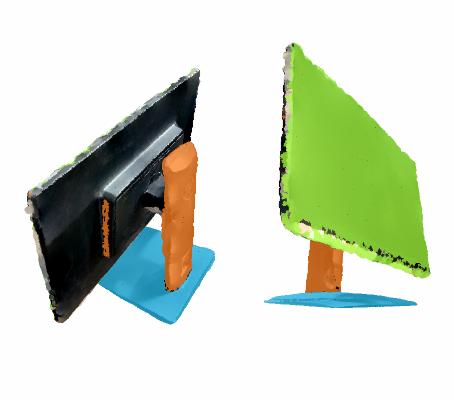} & \includegraphics[width=0.151\textwidth]{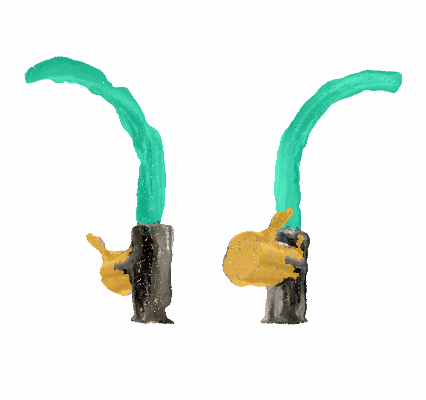} \\
        {\scriptsize\makecell[c]{\projname{} \\(Ours)}} & \includegraphics[width=0.18\textwidth]{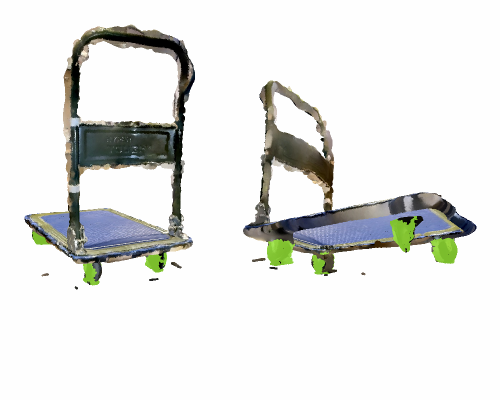} & \includegraphics[width=0.199\textwidth]{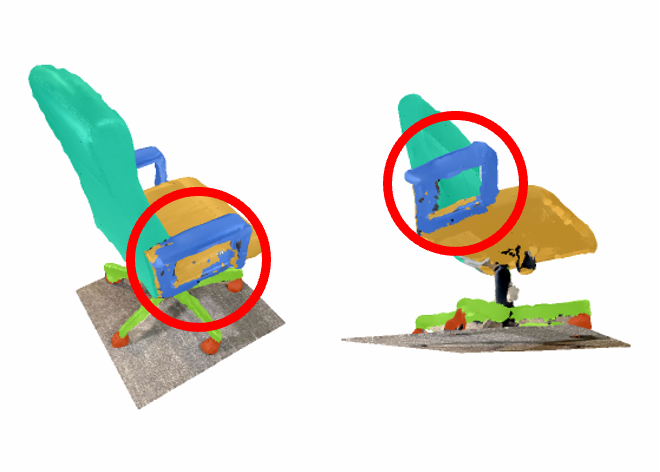} & \includegraphics[width=0.103\textwidth]{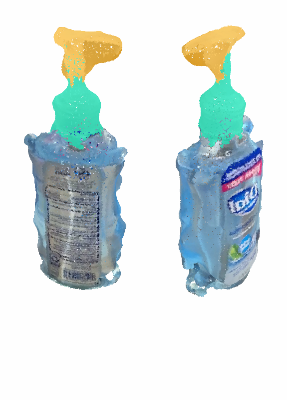} & \includegraphics[width=0.161\textwidth]{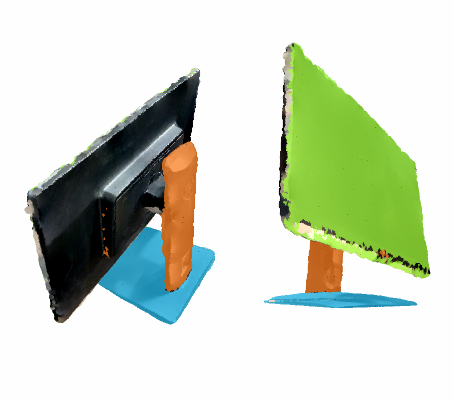} & \includegraphics[width=0.151\textwidth]{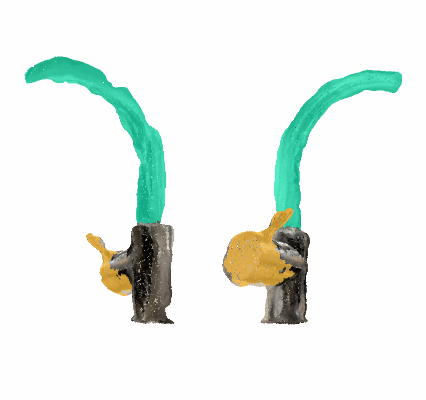} \\

        \end{tabularx}
    }
    \\
    \scriptsize{
        \renewcommand{\arraystretch}{0.0}
        \setlength{\tabcolsep}{0.0em}
        \setlength{\fboxrule}{0.0pt}
        \setlength{\fboxsep}{0pt}
        
        \begin{tabularx}{0.96\textwidth}{>{\centering\arraybackslash}m{0.16\textwidth}|>{\centering\arraybackslash}m{0.148\textwidth}>{\centering\arraybackslash}m{0.209\textwidth}>{\centering\arraybackslash}m{0.196\textwidth}>{\centering\arraybackslash}m{0.09\textwidth}>{\centering\arraybackslash}m{0.150\textwidth}}
        \midrule
        Category & Kettle & KitchenPot & \makecell[c]{Storage\\Furniture} & Suitcase & TrashCan\\[1em]
         
        \makecell[c]{Text Prompt} & 
        {\scriptsize\makecell[c]{\textcolor{color_skyblue}{lid},\textcolor{color_lightgreen}{handle},\\ \textcolor{color_orange}{spout}}} & 
        {\scriptsize\makecell[c]{\textcolor{color_cyan}{lid},\textcolor{color_yellow}{handle}}} & 
        {\scriptsize\makecell[c]{\textcolor{color_skyblue}{door},\textcolor{color_lightgreen}{drawer},\\ \textcolor{color_orange}{handle}}} & 
        {\scriptsize\makecell[c]{\textcolor{color_cyan}{handle},\\ \textcolor{color_yellow}{wheel}}} & 
        {\scriptsize\makecell[c]{\textcolor{color_skyblue}{footpedal},\\ \textcolor{color_lightgreen}{lid},\textcolor{color_orange}{door}}}\\[1em]
        \midrule
        Input & \includegraphics[width=0.148\textwidth]{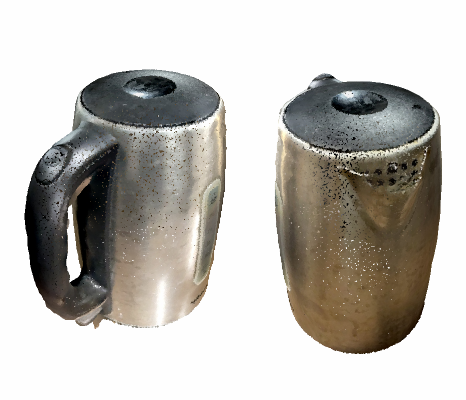} & \includegraphics[width=0.209\textwidth]{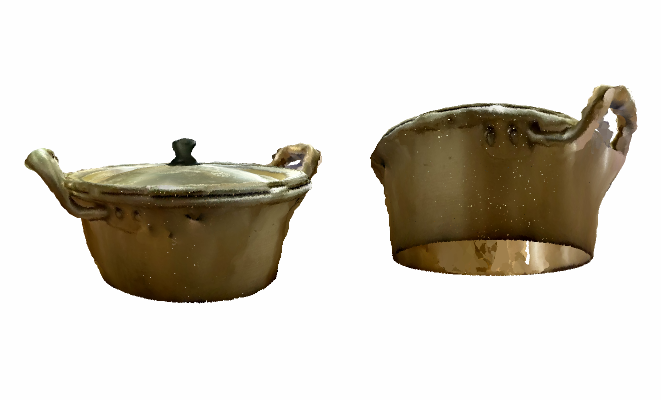} & \includegraphics[width=0.196\textwidth]{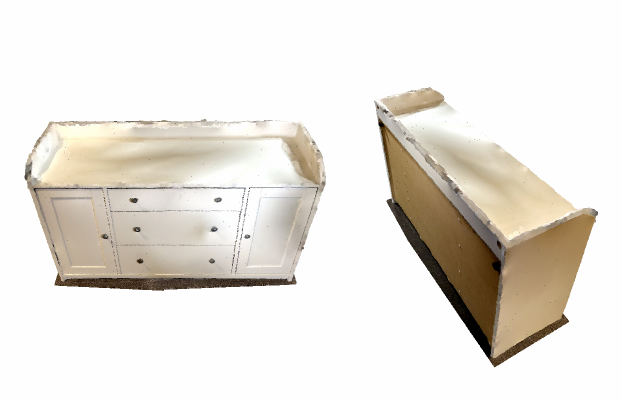} & \includegraphics[width=0.09\textwidth]{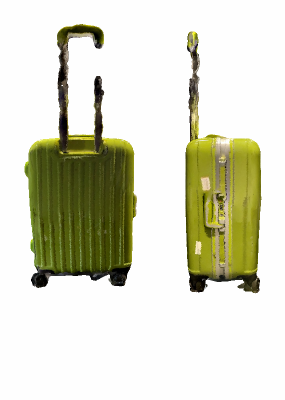} & \includegraphics[width=0.150\textwidth]{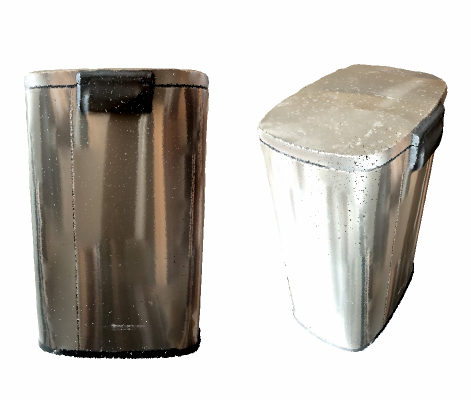}\\
        PartSLIP~\cite{liu2023partslip} & \includegraphics[width=0.148\textwidth]{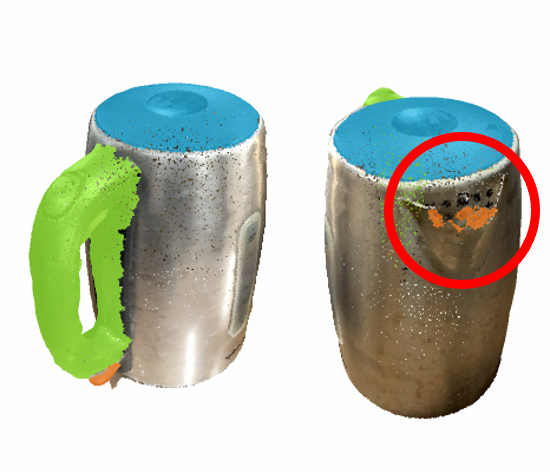} & \includegraphics[width=0.209\textwidth]{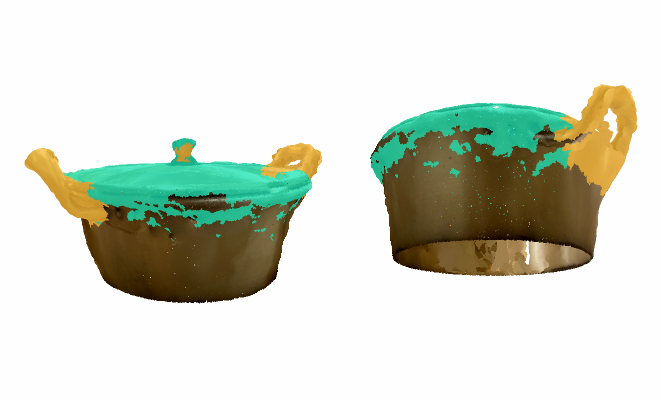} & \includegraphics[width=0.196\textwidth]{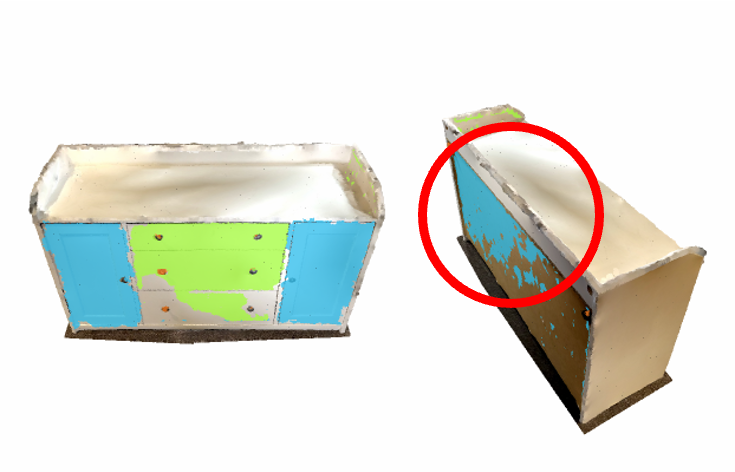} & \includegraphics[width=0.09\textwidth]{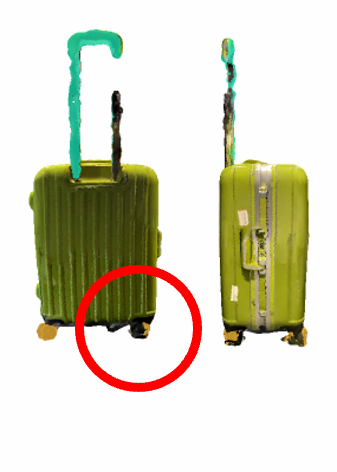} & \includegraphics[width=0.150\textwidth]{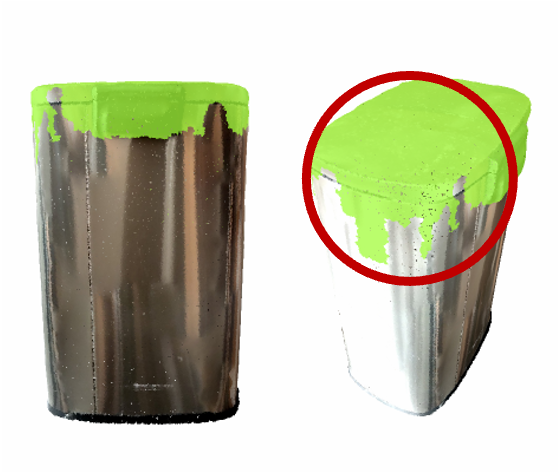}\\
        {\scriptsize\makecell[c]{\projname{} \\(Ours)}} & \includegraphics[width=0.148\textwidth]{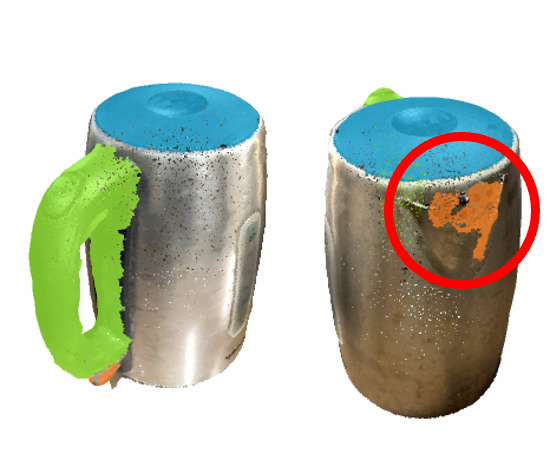} & \includegraphics[width=0.209\textwidth]{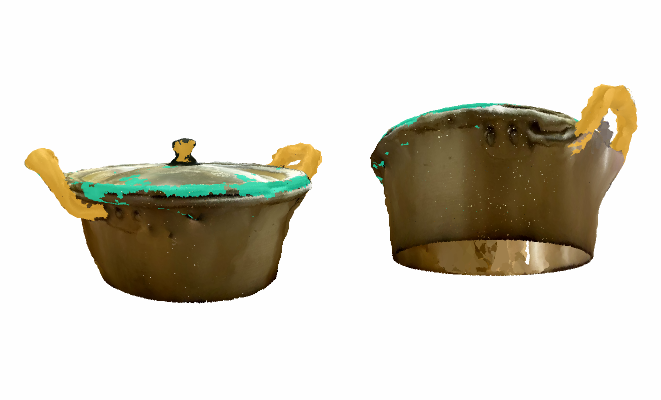} & \includegraphics[width=0.196\textwidth]{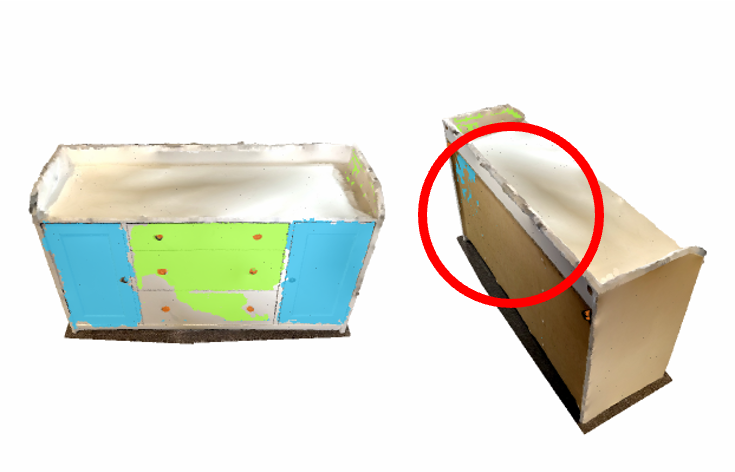} & \includegraphics[width=0.09\textwidth]{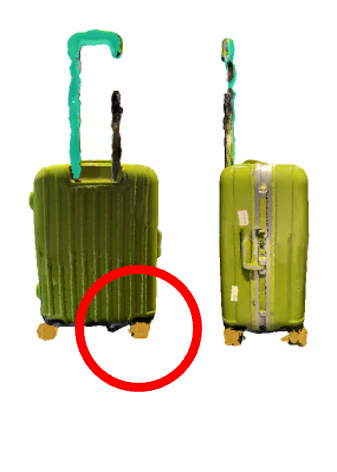} & \includegraphics[width=0.150\textwidth]{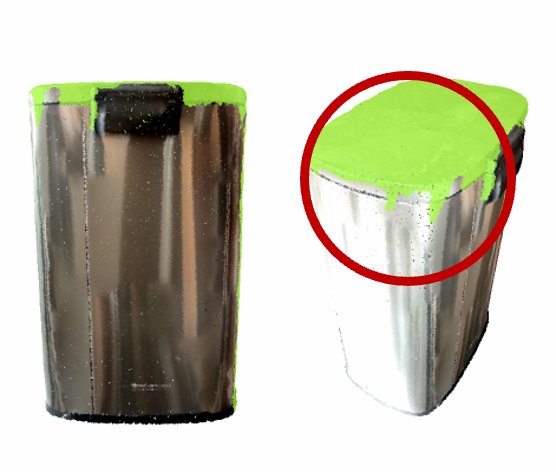}\\

        \end{tabularx}
    }
    \caption{Qualitative results on real-world scan data. In the highlighted red circle, it is evident that our method achieves more accurate segmentation than PartSLIP~\cite{liu2023partslip}.}
    \label{fig:supp_real_pc}
\end{figure*}

\section{Results on Real-World Scanned Data}
\label{sec: Results on Real-world Data}
Fig.~\ref{fig:supp_real_pc} illustrates the results of semantic segmentation on the real-world scan data used in PartSLIP~\cite{liu2023partslip}, which is captured by smartphone. As seen in the figure, our method demonstrates its robustness by successfully predicting not only with higher-quality real-world scans like OmniObject3D~\cite{wu2023omniobject3d} as illustrated in Fig.~\ref{fig:omniobject3d_supp}~\refofpaper{} but also with lower-quality scan data. Also, our method provides more accurate segmentation than PartSLIP~\cite{liu2023partslip}. In the case of the Chair, our method accurately segments the arm part, while PartSLIP fails to do so. For the Kettle, our method better identifies the spout compared to PartSLIP. Additionally, in the cases of StorageFurniture and TrashCan, PartSLIP segments parts that should not be segmented (the backside of StorageFurniture and the lid of TrashCan). On the other hand, for the KitchenPot, while PartSLIP finds the lid part that our method misses, its boundary is still not perfect. We demonstrate that our approach identifies more accurate parts while simultaneously predicting more precise boundaries.

\section{Complete Quantitative Results of \projname{}}
\label{sec: Whole Quantitative Result}
Tab.~\ref{tab:supp_full_semantic}, Tab.~\ref{tab:supp_full_partaware_instance}, and Tab.~\ref{tab:supp_full_partagnostic_instance} show the full table of quantitative results for semantic segmentation, part-aware instance segmentation, and part-agnostic instance segmentation, respectively. Overall, our method demonstrates the best results across whole categories and parts. Please refer to the complete table on the subsequent page for comprehensive information. Moreover, after quantitative result tables, additional qualitative results are illustrated in Sec.~\ref{sec: Additional Qualitative Results Semantic} and Sec.~\ref{sec: Additional Qualitative Results Instance}.

\begin{table*}[t!]
\caption{Full table of semantic segmentation results on the PartNet-Mobility~\cite{Xiang_2020_SAPIEN} dataset.}
\hspace{-2.5\baselineskip}
     {\tiny
    \setlength{\tabcolsep}{0.1em}
    {
\begin{tabularx}{0.57\linewidth}[t]
    {>{\centering\arraybackslash}m{0.08\textwidth}>{\centering\arraybackslash}m{0.08\textwidth}|>{\centering\arraybackslash}m{0.0575\textwidth}>{\centering\arraybackslash}m{0.0575\textwidth}>{\centering\arraybackslash}m{0.08\textwidth}|>{\centering\arraybackslash}m{0.065\textwidth}>{\centering\arraybackslash}m{0.06\textwidth}|>{\centering\arraybackslash}m{0.055\textwidth}|}\toprule
 & &\multicolumn{3}{c|}{\tiny{Baselines}} & \multicolumn{2}{c|}{\tiny{Ablations}}& \\
\midrule
\tiny{Category} & \tiny{Part}&\tiny{SATR}&\tiny{\makecell[c]{SATR\\+SP}}&\tiny{PartSLIP}&\tiny{\makecell[c]{\textit{w/o}\\\textit{Weight}\\\textit{ Pred.}}}&\tiny{\makecell[c]{\textit{w/o}\\\textit{SAM}\\\textit{Integ.}}}& \tiny{\textbf{Ours}}\\
\midrule\midrule
\multirow{1}{*}{Bottle}&lid & 50.0& 55.5& 80.7& \textbf{85.9}& 81.0& 83.6\\
\midrule
\multirow{1}{*}{Box}&lid & 50.6& 54.9& 77.9& \textbf{84.3}& 82.8& 81.1\\
\midrule
\multirow{1}{*}{Bucket}&handle & 6.6& 7.5& 21.1& \textbf{85.4}& 52.4& 83.6\\
\midrule
\multirow{2}{*}{Camera}&button & 10.2& 20.8& 46.1& 48.0& \textbf{53.3}& 51.6\\
 &lens & 24.1& 27.5& \textbf{79.1}& 77.8& 76.6& 77.1\\
\midrule
\multirow{1}{*}{Cart}&wheel & 30.8& 37.2& 78.8& \textbf{85.4}& 81.2& 85.0\\
\bottomrule
    \end{tabularx}} 
    {\begin{tabularx}{0.57\linewidth}[t]
    {>{\centering\arraybackslash}m{0.08\textwidth}>{\centering\arraybackslash}m{0.08\textwidth}|>{\centering\arraybackslash}m{0.0575\textwidth}>{\centering\arraybackslash}m{0.0575\textwidth}>{\centering\arraybackslash}m{0.08\textwidth}|>{\centering\arraybackslash}m{0.065\textwidth}>{\centering\arraybackslash}m{0.06\textwidth}|>{\centering\arraybackslash}m{0.055\textwidth}|}\toprule
 & &\multicolumn{3}{c|}{\tiny{Baselines}} & \multicolumn{2}{c|}{\tiny{Ablations}}& \\
\midrule
\tiny{Category} & \tiny{Part}&\tiny{SATR}&\tiny{\makecell[c]{SATR\\+SP}}&\tiny{PartSLIP}&\tiny{\makecell[c]{\textit{w/o}\\\textit{Weight}\\\textit{ Pred.}}}&\tiny{\makecell[c]{\textit{w/o}\\\textit{SAM}\\\textit{Integ.}}}& \tiny{\textbf{Ours}}\\
\midrule\midrule
\multirow{4}{*}{Microwave}&display & 1.5& 0.0& 28.4& 28.3& 27.9& \textbf{51.0}\\
 &door & 17.7& 16.6& 41.8& 52.2& \textbf{62.2}& 58.8\\
 &handle & 4.5& 25.4& 77.2& 86.9& \textbf{90.0}& 84.9\\
 &button & 0.7& 0.5& 23.6& 25.7& 25.0& \textbf{41.5}\\
 \midrule
\multirow{3}{*}{Mouse}&button & 7.7& 6.7& 16.7& \textbf{21.0}& 8.5& 8.8\\
 &cord & 29.1& 31.0& 63.5& 65.9& \textbf{66.2}& 65.9\\
\bottomrule
    \end{tabularx}}
    }
\label{tab:supp_full_semantic}
\end{table*}

\begin{table*}[t!]
    \caption{Full table of part-aware instance segmentation results on the PartNet-Mobility~\cite{Xiang_2020_SAPIEN} dataset.}
     {\tiny
    \setlength{\tabcolsep}{0.1em}
    {\begin{tabularx}{0.5\linewidth}[t]
    {>{\centering\arraybackslash}m{0.09\textwidth}>{\centering\arraybackslash}m{0.089\textwidth}|>{\centering\arraybackslash}m{0.084\textwidth}|>{\centering\arraybackslash}m{0.07\textwidth}>{\centering\arraybackslash}m{0.07\textwidth}|>{\centering\arraybackslash}m{0.07\textwidth}|}\toprule

 & & \tiny{Baselines} & \multicolumn{2}{c|}{\tiny{Ablations}}& \\
\midrule
Category&Part&PartSLIP&\makecell[c]{\textit{w/o}\\\textit{Weight}\\\textit{Pred.}}&\tiny{\makecell[c]{\textit{w/o}\\\textit{SAM}\\\textit{Integ.}}}&\textbf{Ours}\\[0.2em]
\midrule\midrule
\multirow{1}{*}{Bottle} &\tiny{lid} & 73.4& 74.9& \textbf{79.5}& 77.4\\
\midrule
\multirow{1}{*}{Box} &\tiny{lid} & 56.3& \textbf{72.2}& 60.7& 62.9\\
\midrule
\multirow{1}{*}{Bucket} &\tiny{handle} & 10.2& \textbf{78.2}& 39.3& 78.2\\
\midrule
\multirow{2}{*}{Camera} &\tiny{button} & 37.6& 36.7& \textbf{39.7}& 38.7\\
 &\tiny{lens} & \textbf{35.8}& 32.7& 29.6& 29.8\\
\midrule
\multirow{1}{*}{Cart} &\tiny{wheel} & 69.5& \textbf{74.0}& 68.3& 71.6\\
\bottomrule
    \end{tabularx}} 
    {\begin{tabularx}{0.5\linewidth}[t]
    {>{\centering\arraybackslash}m{0.09\textwidth}>{\centering\arraybackslash}m{0.089\textwidth}|>{\centering\arraybackslash}m{0.084\textwidth}|>{\centering\arraybackslash}m{0.07\textwidth}>{\centering\arraybackslash}m{0.07\textwidth}|>{\centering\arraybackslash}m{0.07\textwidth}|}\toprule

 & & \tiny{Baselines} & \multicolumn{2}{c|}{\tiny{Ablations}}& \\
\midrule
Category&Part&PartSLIP&\makecell[c]{\textit{w/o}\\\textit{Weight}\\\textit{Pred.}}&\tiny{\makecell[c]{\textit{w/o}\\\textit{SAM}\\\textit{Integ.}}}&\textbf{Ours}\\[0.2em]
\midrule\midrule
\multirow{4}{*}{Microwave} &\tiny{display} & 33.7& 33.7& 33.7& \textbf{38.1}\\
 &\tiny{door} & 39.0& 23.2& \textbf{42.7}& 27.2\\
 &\tiny{handle} & 50.5& 50.5& \textbf{60.4}& 50.5\\
 &\tiny{button} & 12.0& 12.0& 12.9& \textbf{13.6}\\
\midrule
\multirow{3}{*}{Mouse} &\tiny{button} & \textbf{5.0}& 3.0& 5.0& 5.0\\
 &\tiny{cord} & \textbf{66.3}& 66.3& 66.3& 66.3\\
\bottomrule
    \end{tabularx}}
}
    \label{tab:supp_full_partaware_instance}
\end{table*}

\begin{table*}[t!]
\caption{Full table of part-agnostic instance segmentation results on the PartNet-Mobility~\cite{Xiang_2020_SAPIEN} dataset.}
     {\tiny
    \setlength{\tabcolsep}{0.1em}
    {\begin{tabularx}{0.5\linewidth}[t]
    {>{\centering\arraybackslash}m{0.10\textwidth}|>{\centering\arraybackslash}m{0.072\textwidth}>{\centering\arraybackslash}m{0.09\textwidth}|>{\centering\arraybackslash}m{0.07\textwidth}>{\centering\arraybackslash}m{0.07\textwidth}|>{\centering\arraybackslash}m{0.07\textwidth}|}\toprule

 & \multicolumn{2}{c|}{\tiny{Baselines}} & \multicolumn{2}{c|}{\tiny{Ablations}}& \\
\midrule
Category&SAM3D&PartSLIP&\makecell[c]{\textit{w/o}\\\textit{Weight}\\\textit{Pred.}}&\tiny{\makecell[c]{\textit{w/o}\\\textit{SAM}\\\textit{Integ.}}}&\textbf{Ours}\\[0.2em]
\midrule\midrule
Bottle& 15.3& 73.4& 74.9& \textbf{79.5}& 76.2\\
\midrule
Box& 18.2& 56.3& \textbf{72.2}& 60.7& 69.6\\
\midrule
Bucket& 12.0& 10.2& \textbf{78.2}& 39.3& 78.2\\
\midrule
Camera& 2.2& 37.1& 35.5& \textbf{37.8}& 37.0\\
\midrule
Cart& 8.5& 69.5& \textbf{74.0}& 68.3& 71.2\\
\midrule
Chair& 5.7& 80.7& 81.1& 81.2& \textbf{83.3}\\
\midrule
Clock& 1.0& 18.7& \textbf{32.3}& 18.6& 18.5\\
\midrule
\makecell[c]{Coffee\\Machine}& 3.6& \textbf{9.8}& 9.5& 9.4& 9.4\\
\midrule
Dishwasher& 6.7& 39.0& 41.9& 47.1& \textbf{49.7}\\
\midrule
Dispenser& 20.9& 59.0& 59.9& \textbf{69.0}& 65.5\\
\midrule
Display& 34.1& 61.5& 66.4& 66.0& \textbf{68.4}\\
\midrule
Door& \textbf{17.2}& 11.5& 11.6& 13.9& 16.0\\
\midrule
Eyeglasses& 27.6& 71.8& 72.1& 72.6& \textbf{73.9}\\
\midrule
Faucet& 23.7& \textbf{42.8}& 37.1& 37.9& 35.6\\
\midrule
\makecell[c]{Folding\\Chair}& 27.2& 86.4& 89.6& \textbf{100.0}& 91.3\\
\midrule
Globe& 4.2& 92.1& \textbf{100.0}& 98.8& 84.3\\
\midrule
Kettle& 41.8& 69.5& 75.3& \textbf{76.3}& 75.9\\
\midrule
Keyboard& 1.0& 34.1& 31.0& 47.3& \textbf{48.9}\\
\midrule
KitchenPot& 31.9& 55.4& \textbf{66.2}& 52.3& 65.4\\
\midrule
Knife& 2.2& \textbf{44.5}& 41.3& 39.2& 32.1\\
\midrule
Lamp& 48.4& 77.0& \textbf{77.7}& 66.5& 75.9\\
\midrule
Laptop& 1.9& 19.5& 26.5& 27.4& \textbf{34.1}\\
\midrule
Lighter& 8.9& 32.7& \textbf{49.9}& 26.6& 39.8\\
\bottomrule
    \end{tabularx}}
    {\begin{tabularx}{0.5\linewidth}[t]
    {>{\centering\arraybackslash}m{0.10\textwidth}|>{\centering\arraybackslash}m{0.072\textwidth}>{\centering\arraybackslash}m{0.09\textwidth}|>{\centering\arraybackslash}m{0.07\textwidth}>{\centering\arraybackslash}m{0.07\textwidth}|>{\centering\arraybackslash}m{0.07\textwidth}|}\toprule

 & \multicolumn{2}{c|}{\tiny{Baselines}} & \multicolumn{2}{c|}{\tiny{Ablations}}& \\
\midrule
Category&SAM3D&PartSLIP&\makecell[c]{\textit{w/o}\\\textit{Weight}\\\textit{Pred.}}&\tiny{\makecell[c]{\textit{w/o}\\\textit{SAM}\\\textit{Integ.}}}&\textbf{Ours}\\[0.2em]
\midrule\midrule
Microwave& 1.4& 15.0& 12.9& \textbf{18.7}& 16.2\\
\midrule
Mouse& 2.5& 12.9& 11.2& 20.5& \textbf{21.0}\\
\midrule
Oven& 1.0& 46.8& 55.6& 56.4& \textbf{62.0}\\
\midrule
Pen& 8.2& 40.8& 35.8& \textbf{47.2}& 34.5\\
\midrule
Phone& 1.1& 31.0& 33.3& 31.6& \textbf{34.7}\\
\midrule
Pliers& 22.2& 3.2& 4.1& \textbf{31.1}& 31.1\\
\bottomrule
    \end{tabularx}
    
    } 
    
}
    \label{tab:supp_full_partagnostic_instance}
\end{table*}

\clearpage
\onecolumn
\section{Additional Qualitative Results for Semantic Segmentation}
\label{sec: Additional Qualitative Results Semantic}
\begin{flushleft}
We present additional part segmentation results below, encompassing semantic segmentation. Each row shows the same object with different views.
\end{flushleft}
\setlength\LTleft{-4mm}
\setlength{\tabcolsep}{0em}
\def\arraystretch{0.0}
\newcolumntype{Z}{>{\centering\arraybackslash}m{0.09\textwidth}}
{\tiny
\centering

\begin{longtable*}{ZZZZZZ|ZZZZZZ}
\toprule
\multicolumn{6}{c|}{view 1}&\multicolumn{6}{c}{view 2}\\
\midrule
Input &
GT &
SATR&
\makecell[c]{SATR\\+SP} &
PartSLIP&
\textbf{Ours}&
Input &
GT &
SATR&
\makecell[c]{SATR\\+SP} &
PartSLIP&
\textbf{Ours}\\

  \midrule
  \endhead

  \bottomrule
  \endfoot

 \multicolumn{6}{c|}{\includegraphics[width=0.54\textwidth]{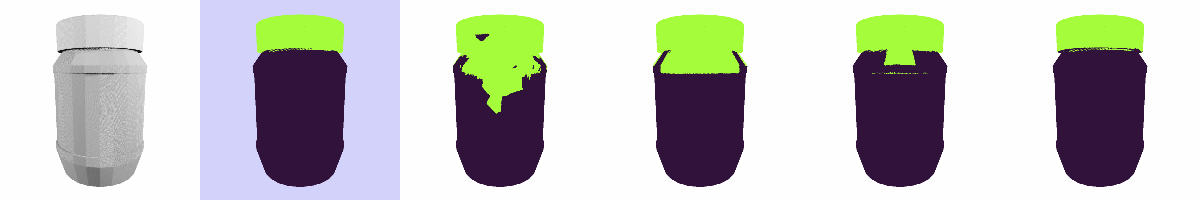}}&\multicolumn{6}{c}{\includegraphics[width=0.54\textwidth]{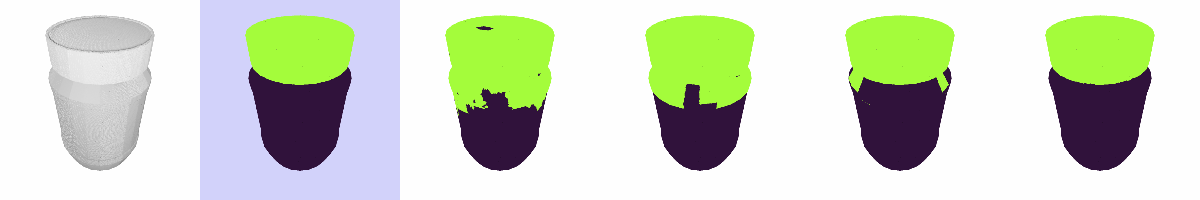}}\\
\multicolumn{6}{c|}{\includegraphics[width=0.54\textwidth]{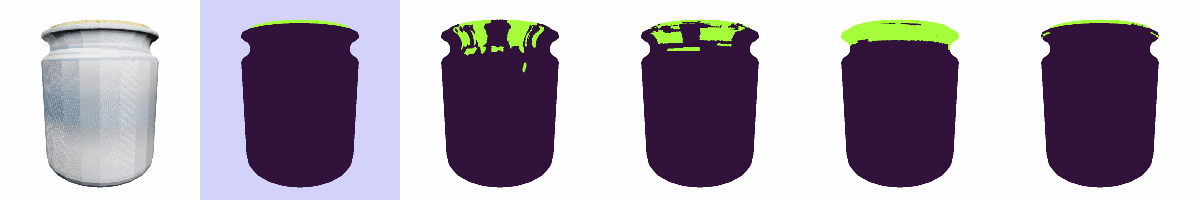}}&\multicolumn{6}{c}{\includegraphics[width=0.54\textwidth]{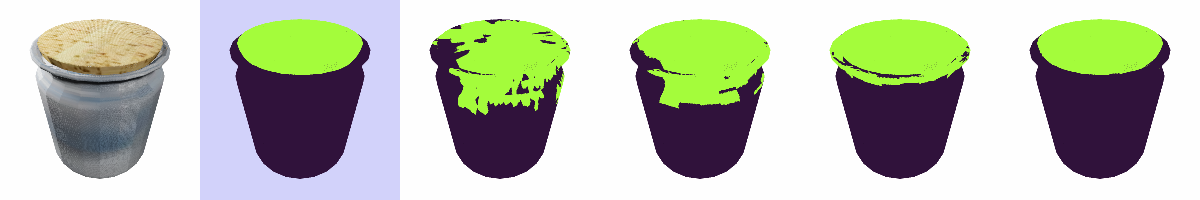}}\\
\multicolumn{6}{c|}{\includegraphics[width=0.54\textwidth]{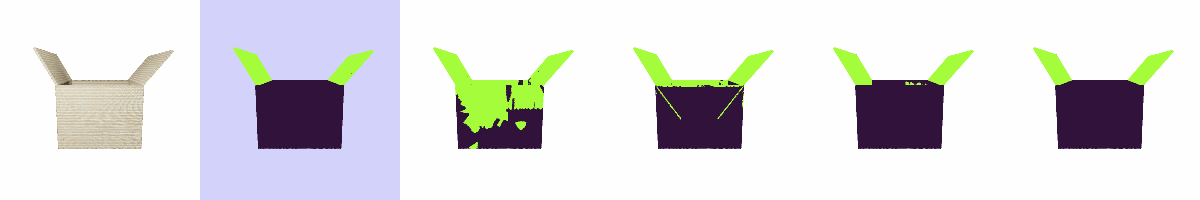}}&\multicolumn{6}{c}{\includegraphics[width=0.54\textwidth]{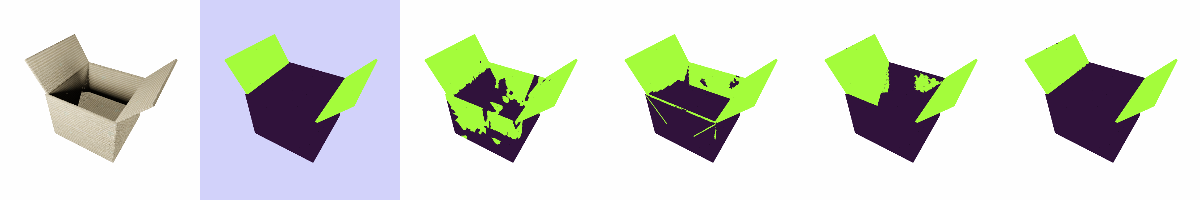}}\\
\multicolumn{6}{c|}{\includegraphics[width=0.54\textwidth]{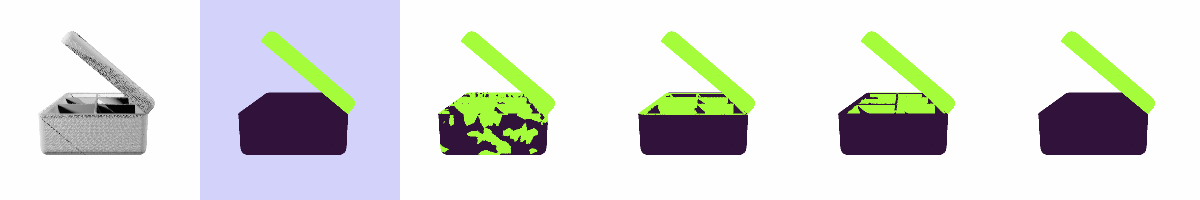}}&\multicolumn{6}{c}{\includegraphics[width=0.54\textwidth]{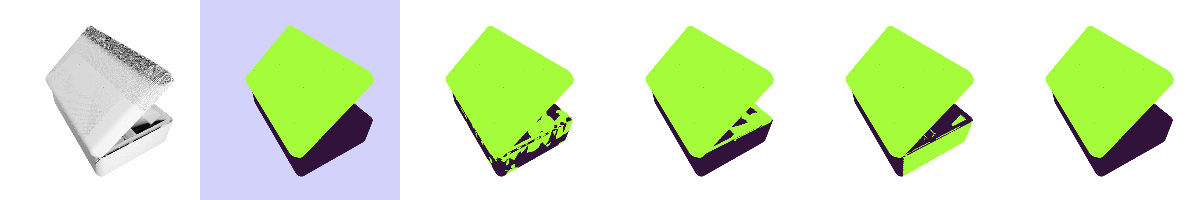}}\\
\multicolumn{6}{c|}{\includegraphics[width=0.54\textwidth]{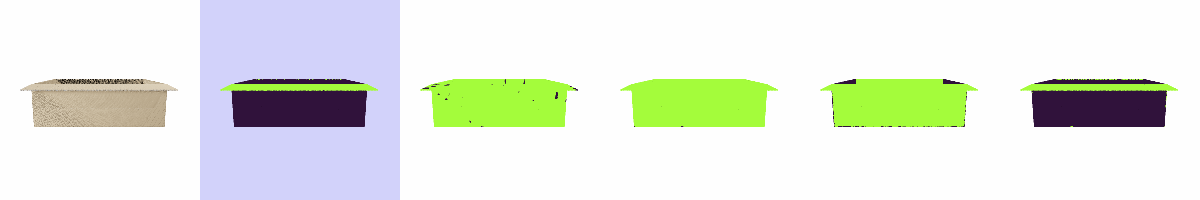}}&\multicolumn{6}{c}{\includegraphics[width=0.54\textwidth]{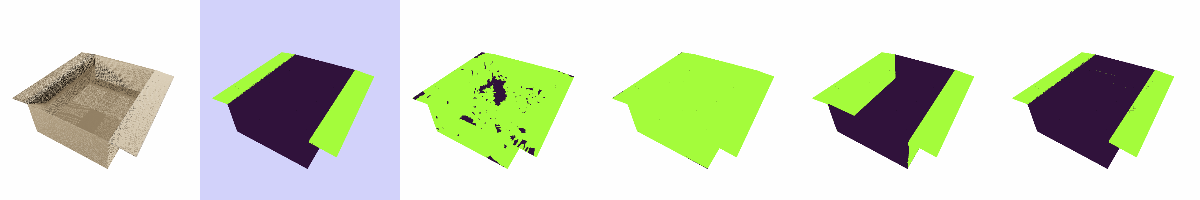}}\\
\multicolumn{6}{c|}{\includegraphics[width=0.54\textwidth]{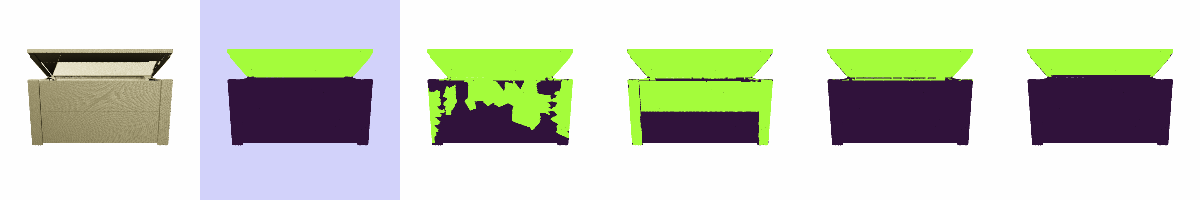}}&\multicolumn{6}{c}{\includegraphics[width=0.54\textwidth]{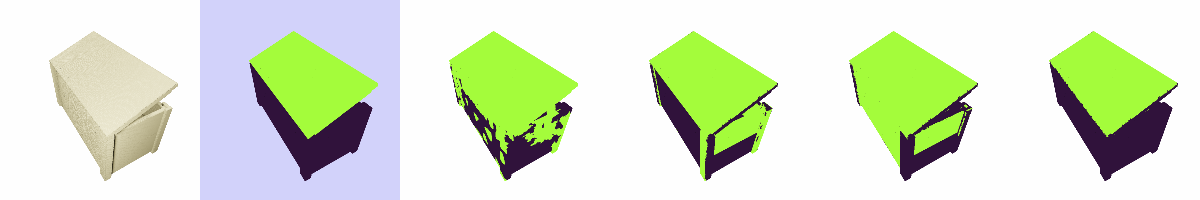}}\\
\multicolumn{6}{c|}{\includegraphics[width=0.54\textwidth]{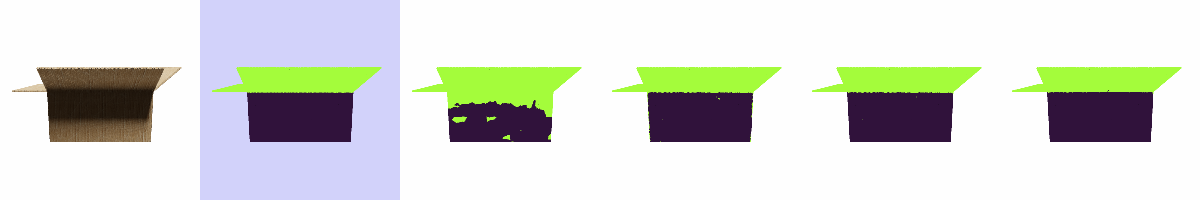}}&\multicolumn{6}{c}{\includegraphics[width=0.54\textwidth]{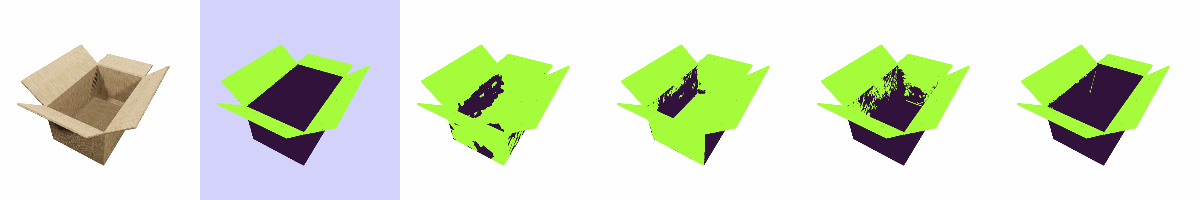}}\\
\multicolumn{6}{c|}{\includegraphics[width=0.54\textwidth]{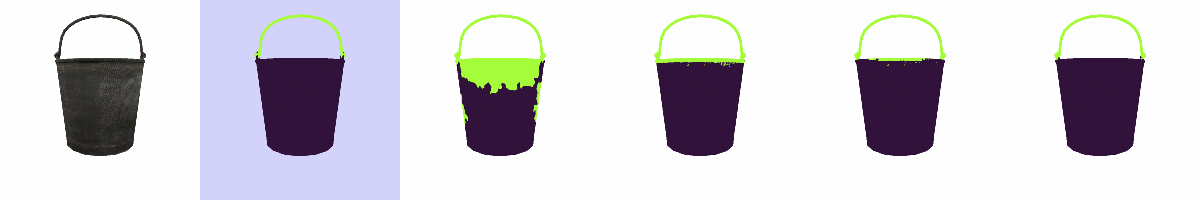}}&\multicolumn{6}{c}{\includegraphics[width=0.54\textwidth]{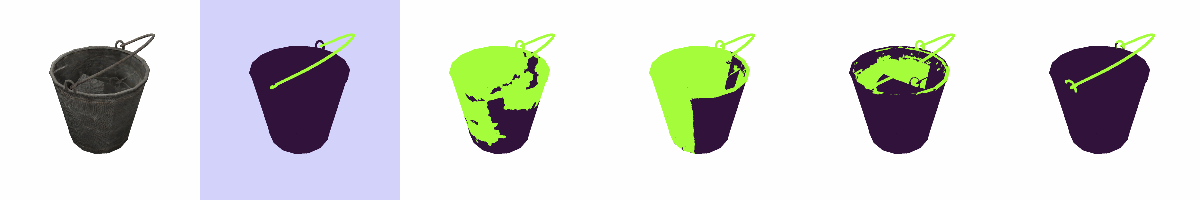}}\\
\end{longtable*}
}

\clearpage
\section{Additional Qualitative Results for Instance Segmentation}
\label{sec: Additional Qualitative Results Instance}
\begin{flushleft}
We present additional part segmentation results below, encompassing instance segmentation. Each row shows the same object with different views.
\end{flushleft}
\setlength{\LTleft}{0mm}
\setlength{\tabcolsep}{0em}
\def\arraystretch{0.0}
\newcolumntype{Z}{>{\centering\arraybackslash}m{0.1\textwidth}}
{\tiny
\centering
\begin{longtable*}{ZZZZZ|ZZZZZ}
\toprule
\multicolumn{5}{c|}{view 1}&\multicolumn{5}{c}{view 2}\\
\midrule
Input &
GT &
SAM3D&
PartSLIP&
\textbf{Ours} &
Input &
GT &
SAM3D&
PartSLIP&
\textbf{Ours}\\

  \midrule
  \endhead

  \bottomrule
  \endfoot

 \multicolumn{5}{c|}{\includegraphics[width=0.5\textwidth]{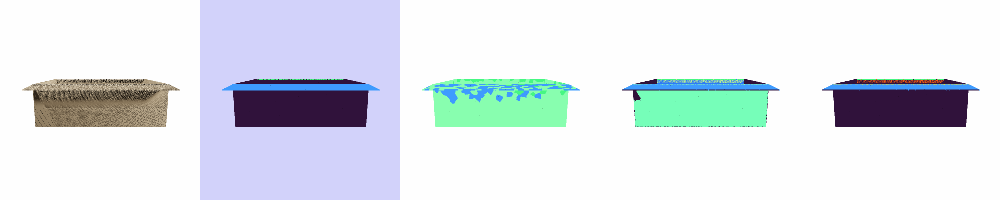}}&\multicolumn{5}{c}{\includegraphics[width=0.5\textwidth]{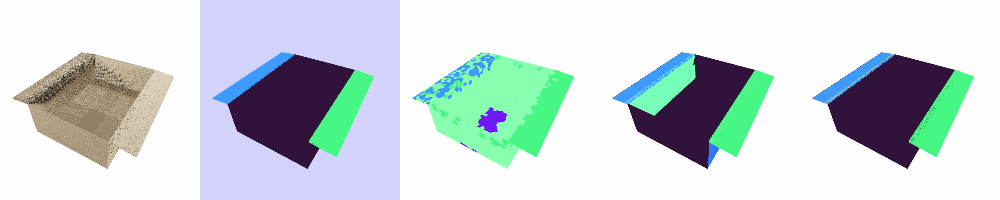}}\\
\multicolumn{5}{c|}{\includegraphics[width=0.5\textwidth]{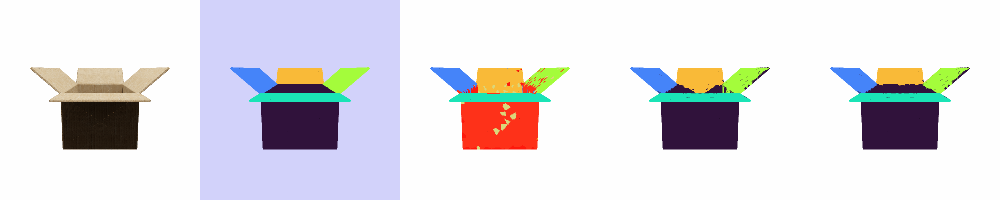}}&\multicolumn{5}{c}{\includegraphics[width=0.5\textwidth]{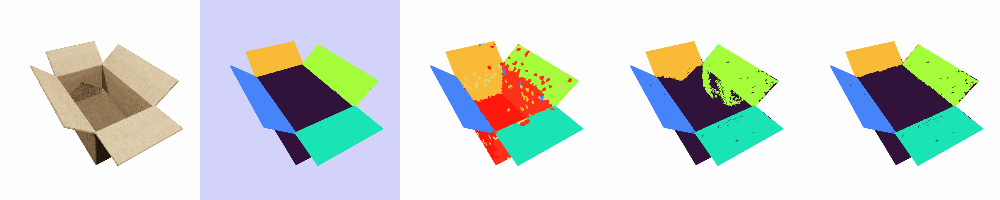}}\\
\multicolumn{5}{c|}{\includegraphics[width=0.5\textwidth]{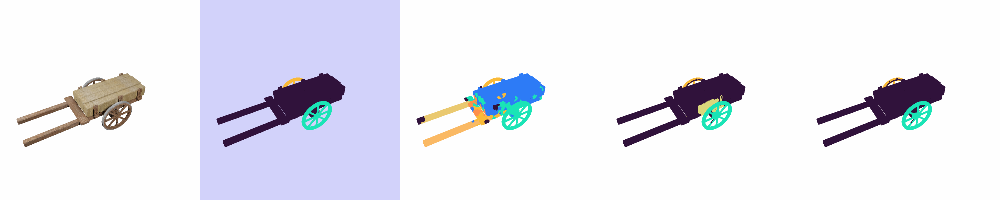}}&\multicolumn{5}{c}{\includegraphics[width=0.5\textwidth]{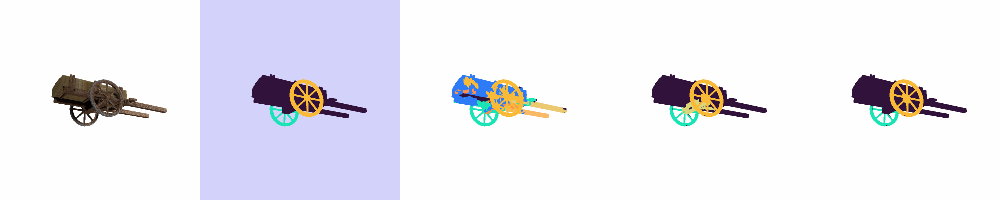}}\\
\multicolumn{5}{c|}{\includegraphics[width=0.5\textwidth]{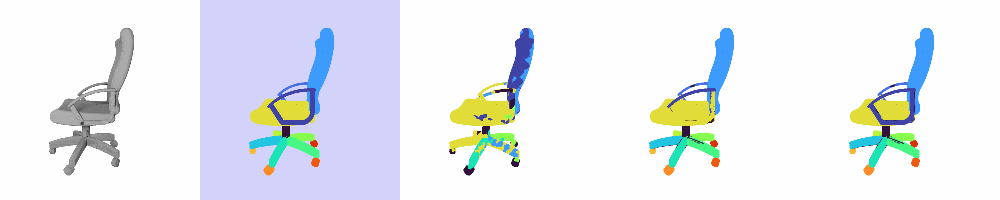}}&\multicolumn{5}{c}{\includegraphics[width=0.5\textwidth]{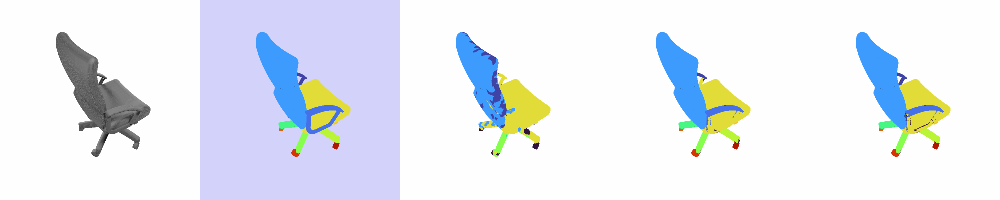}}\\
\multicolumn{5}{c|}{\includegraphics[width=0.5\textwidth]{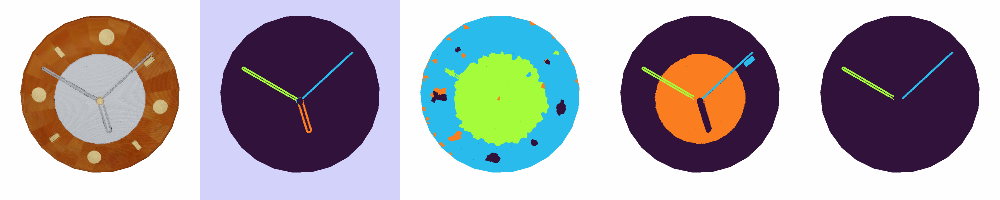}}&\multicolumn{5}{c}{\includegraphics[width=0.5\textwidth]{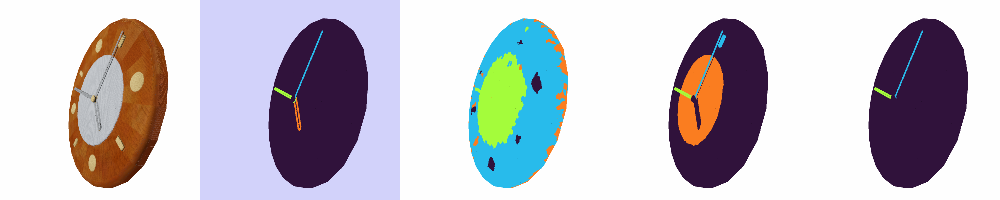}}\\
\multicolumn{5}{c|}{\includegraphics[width=0.5\textwidth]{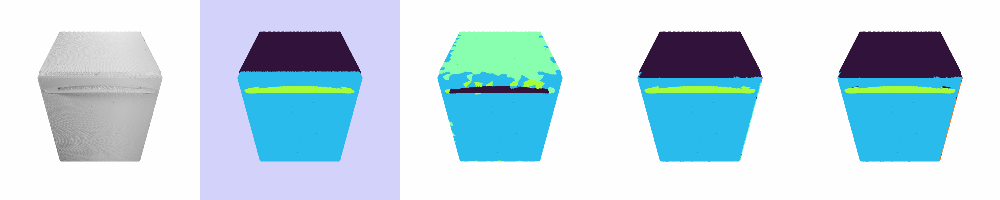}}&\multicolumn{5}{c}{\includegraphics[width=0.5\textwidth]{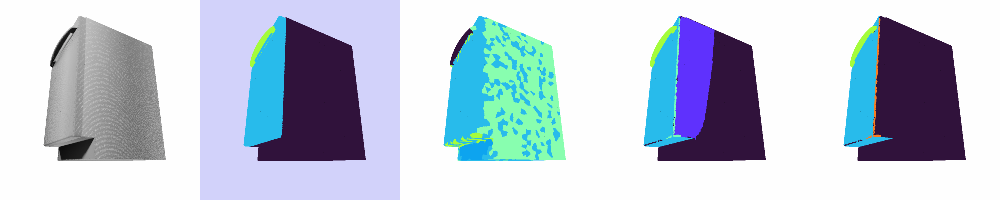}}\\
\multicolumn{5}{c|}{\includegraphics[width=0.5\textwidth]{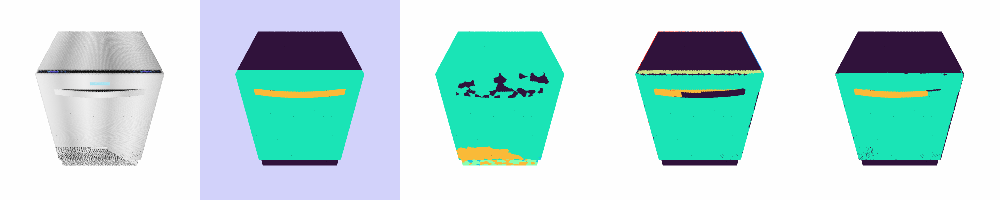}}&\multicolumn{5}{c}{\includegraphics[width=0.5\textwidth]{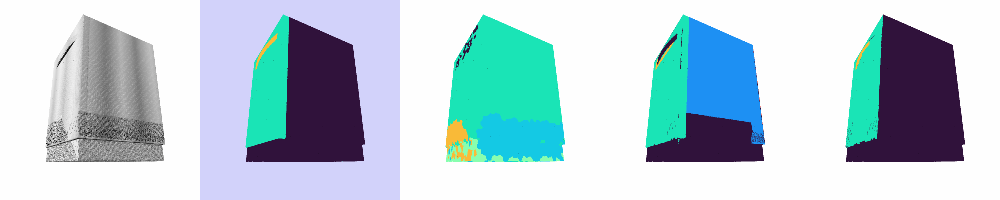}}\\
\multicolumn{5}{c|}{\includegraphics[width=0.5\textwidth]{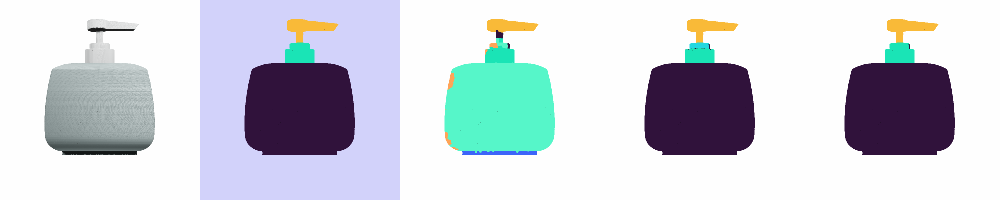}}&\multicolumn{5}{c}{\includegraphics[width=0.5\textwidth]{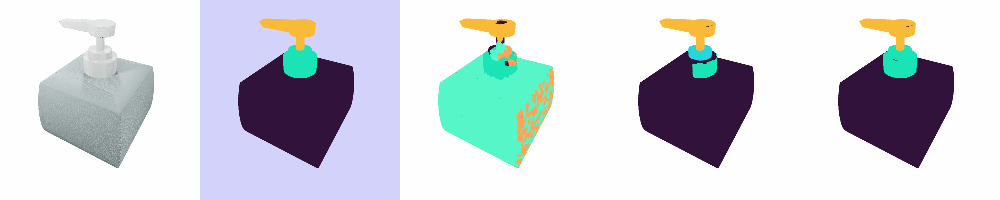}}\\
\end{longtable*}
}

\clearpage

\end{document}